\documentclass{article}
\usepackage{iclr2027_conference,times}

\usepackage{amsmath,amsfonts,bm}

\def\eqref#1{equation~\ref{#1}}

\def\1{\bm{1}}

\DeclareMathAlphabet{\mathsfit}{\encodingdefault}{\sfdefault}{m}{sl}
\SetMathAlphabet{\mathsfit}{bold}{\encodingdefault}{\sfdefault}{bx}{n}

\usepackage{booktabs}
\usepackage{longtable}
\usepackage{array}
\usepackage{graphicx}
\usepackage{adjustbox}
\usepackage{multirow}
\usepackage{pifont}
\usepackage{fontawesome5}
\usepackage{makecell}
\usepackage[most]{tcolorbox}
\usepackage[font=footnotesize,skip=3pt,subrefformat=parens]{subcaption}
\usepackage{placeins}
\usepackage[table]{xcolor}
\usepackage{hyperref}
\usepackage{url}

\title{\textsc{CoVLM-Bench}: A Real-World Benchmark\\
for Cooperative Driving\\
Question Answering and Planning}

\author{Kang Yang$^{1}$, \enspace Shuai Liu$^{1}$, \enspace Hang Li$^{1}$, \enspace Yance Fang$^{1}$, \enspace Deying Li$^{1,2}$, \enspace Yongcai Wang$^{1,2,*}$ \\
\normalfont $^{1}$Renmin University of China \\
\normalfont $^{2}$Hebei Key Laboratory of Real-virtual Integrated Autonomous Systems (RIAS) \\[0.2em]
\normalfont\small\texttt{y1127238112@gmail.com}, \enspace \texttt{rucliushuai@ruc.edu.cn}, \enspace \texttt{lhruc@ruc.edu.cn} \\
\normalfont\small\texttt{fyc@ruc.edu.cn}, \enspace \texttt{deyingli@ruc.edu.cn}, \enspace \texttt{ycw@ruc.edu.cn} \\
\normalfont\small $^{*}$Corresponding author \\[0.35em]
\normalfont\small Code: \url{https://github.com/sidiangongyuan/CoVLM-Bench} \\
\normalfont\small Project page: \url{https://sidiangongyuan.github.io/CoVLM-Bench/}}

\newcommand{\dataset}{\textsc{CoVLM-Bench}}
\newcommand{\model}{\textsc{CoVLM-Drive}}

\definecolor{egoBlue}{RGB}{55,126,184}
\definecolor{infraOrange}{RGB}{255,127,0}
\definecolor{bothGreen}{RGB}{77,175,74}
\definecolor{unkGray}{RGB}{150,150,150}
\definecolor{benchBlue}{RGB}{37,99,235}
\definecolor{benchRed}{RGB}{192,0,0}
\definecolor{benchGreen}{RGB}{0,176,80}
\definecolor{benchAmber}{RGB}{245,158,11}
\definecolor{benchGray}{RGB}{94,106,125}
\definecolor{tableBandGray}{RGB}{238,239,241}
\definecolor{tableBandBlue}{RGB}{226,238,249}
\definecolor{tableFocusBlue}{RGB}{219,234,254}
\definecolor{tableVariantGray}{RGB}{229,231,235}
\definecolor{tableGoodGreen}{RGB}{231,244,235}
\definecolor{tableBadRed}{RGB}{249,234,232}
\definecolor{tableNeutralGray}{RGB}{243,244,246}
\definecolor{tableRefBlue}{RGB}{235,242,249}
\hypersetup{
  colorlinks=true,
  linkcolor=benchBlue,
  citecolor=benchBlue,
  urlcolor=benchBlue
}
\newcommand{\greenword}{\textcolor{benchGreen!60!black}{green}}
\newcommand{\redword}{\textcolor{benchRed!85!black}{red}}
\newcommand{\amberword}{\textcolor{benchAmber!80!black}{amber}}
\newcommand{\blueword}{\textcolor{benchBlue}{blue}}
\newcommand{\grayword}{\textcolor{benchGray}{gray}}
\newcommand{\Greenword}{\textcolor{benchGreen!60!black}{Green}}
\newcommand{\cmark}{\textcolor{benchGreen}{\ding{51}}}
\newcommand{\xmark}{\textcolor{benchRed}{\ding{55}}}
\newcommand{\pmark}{\textcolor{benchAmber}{\faIcon{adjust}}}
\newcommand{\benchicon}[2]{\textcolor{#1}{\faIcon{#2}}}
\newcommand{\benchref}[1]{\textcolor{benchGray}{\scriptsize(\citeyear{#1})}}
\newcolumntype{M}{>{\centering\arraybackslash}p{3.85em}}
\newcolumntype{Q}{>{\centering\arraybackslash}p{3.40em}}

\newcommand{\qagood}[1]{\cellcolor{tableGoodGreen}\textbf{#1}}
\newcommand{\qabad}[1]{\cellcolor{tableBadRed}#1}

\newcommand{\qactrl}[2]{#1\,{\fontsize{6}{6.3}\selectfont\textcolor{benchGray}{(#2)}}}
\newcommand{\bestcell}[1]{\textbf{#1}}
\newcommand{\runnerup}[1]{\underline{#1}}
\newcommand{\tablefocus}{\rowcolor{tableFocusBlue}}
\newcommand{\tablegroupgray}[2]{%
  \rowcolor{tableBandGray}\multicolumn{#1}{l}{\textbf{#2}}\\}
\newcommand{\tablegroupblue}[2]{%
  \rowcolor{tableBandBlue}\multicolumn{#1}{l}{\textbf{#2}}\\}


\iclrfinalcopy
\begin{document}

\maketitle
\lhead{Preprint}

\begin{abstract}
Vision--language models (VLMs) have made substantial progress in autonomous
driving, but their success has primarily been studied in ego-centric scenes.
Infrastructure-side observations provide views beyond the ego vehicle's field
of view, yet conventional cooperative-driving systems typically transform them
into geometric representations for downstream perception and planning.
Directly incorporating these views into VLMs offers an opportunity to improve
cooperative scene understanding and trajectory planning. However, question
answering and trajectory planning have not been jointly evaluated on the same
real-world vehicle--infrastructure scenes. We present \dataset{}, a benchmark
for cooperative driving question answering (CDQA) and cooperative planning
(CP) on vehicle--infrastructure paired scenes. \dataset{} provides
scene-grounded CDQA annotations, three-part rationales as auxiliary
supervision, and future trajectory targets derived from recorded ego motion.
It contains 2,196 paired frames with 35,136 CDQA annotations, while CP
predicts six waypoints over a three-second horizon. The annotations combine
model-assisted drafting, record-based computation, and human verification.
Built upon \dataset{}, we introduce \model{}, a unified VLM baseline that
directly uses paired views for both CDQA and CP. Experiments show that CDQA
adaptation improves answer accuracy and that \model{} reaches a lower FDE than
the compared V2X planners; QA initialization and rationale supervision each
reduce planning error. Together, \dataset{} and \model{} support the training
and comparison of VLMs for cooperative scene understanding and planning.
\end{abstract}

\section{Introduction}

\begin{figure}[t!]
    \centering
    \includegraphics[width=\linewidth]{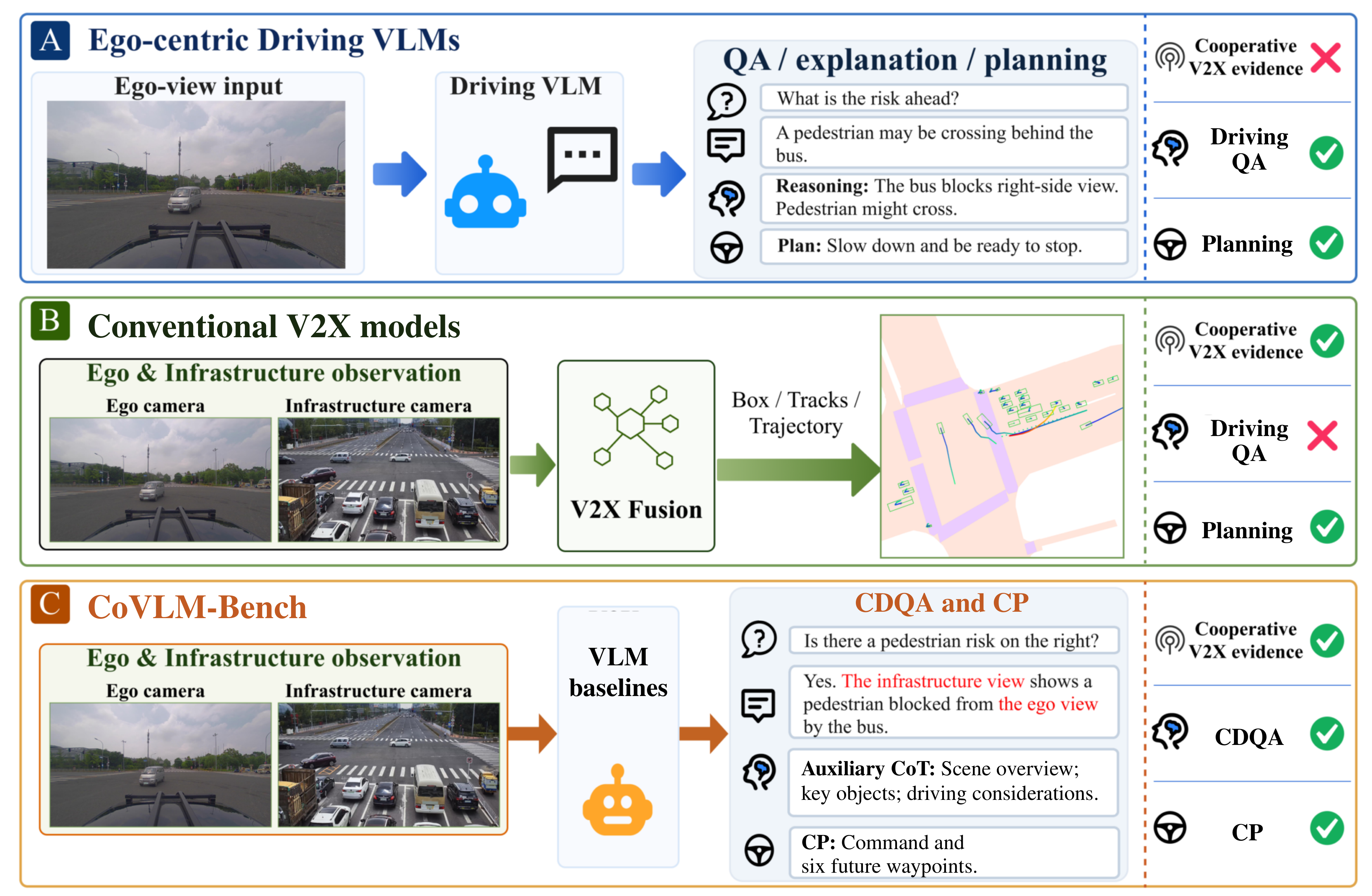}
\caption{\textbf{CDQA and CP in cooperative driving}. The schematic contrasts
ego-centric driving VLMs (A), conventional V2X perception and planning (B), and
the paired-view tasks supported by \dataset{} (C). The benchmark associates
CDQA annotations and CP targets with the same real-world scenes;
\model{} provides a unified VLM baseline for CDQA and CP. Illustrative task
examples are shown.}
\label{fig:teaser}
\end{figure}

Vision--language models (VLMs) have made substantial progress in autonomous
driving, enabling a single model to answer questions about driving scenarios
and predict subsequent vehicle
maneuvers~\citep{drivelm,wang2025omnidrive,hwang2024emma,jiang2024senna}.
These capabilities connect scene understanding to planning: (1) question
answering (QA) offers a measurable assessment of scene
understanding~\citep{nuscenesqa,lingoqa,reason2drive,xie2025drivebench}; (2)
step-by-step chain-of-thought (CoT) rationales express intermediate reasoning
about the scene and provide auxiliary
supervision~\citep{jiang2025alphadrive,fu2025orion,zhou2025autovla,li2025recogdrive,tian2026ccot};
and (3) trajectory planning predicts future ego motion from the observed
scene. However, these capabilities have predominantly been studied with
ego-centric camera inputs.

In cooperative perception, infrastructure-mounted cameras capture traffic
participants outside the ego vehicle's field of view or occluded by other road
users~\citep{dair-v2x,v2x-seq,collaborative-survey2}. Conventional
cooperative-driving pipelines convert these observations into geometric
representations---such as bounding boxes, object tracks, and forecast
states---and pass them to downstream
planners~\citep{where2comm,v2x-vit,yu2025univ2x,song2025unimmv2x,yin2025map}.
These pipelines demonstrate the value of roadside observations for geometric
perception, motivating us to explore how VLMs can use the paired images
directly for scene understanding and trajectory planning.

Studying this opportunity calls for data that support both understanding and
planning on cooperative scenes. V2X-QA compares cooperative question answering
across ego, infrastructure, and paired views, while D2-V2X evaluates grounded
question--rationale--answer outputs for ego-occluded
objects~\citep{you2026v2xqa,richard2026d2v2x}. In parallel,
cooperative-driving datasets provide paired observations and recorded ego
motion for
planning~\citep{dair-v2x,v2x-seq,yu2025univ2x,song2025unimmv2x,yin2025map}.
These resources provide complementary ingredients, but question answering and
trajectory planning have not been jointly evaluated on the same real-world
paired scenes. Bringing these annotations together would allow us to evaluate
both abilities on corresponding scenes and examine whether
understanding-oriented supervision also improves planning.

\dataset{} is a real-world benchmark built from V2X-Seq-SPD's synchronized
vehicle--infrastructure images, cooperative object records, and ego-motion
logs. It supports two tasks on the same paired scenes: cooperative driving
question answering (CDQA) and cooperative planning (CP). CDQA provides
scene-supported answers and, for object-related questions, references to the
supporting participants and the views in which they are visible. CP uses the
recorded ego motion to derive a driving command and six ego-relative waypoints
over a three-second horizon. Complete-horizon planning examples also include a
three-part rationale as auxiliary supervision.
Qwen3-VL-32B~\citep{bai2025qwen3vl} drafts the textual descriptions, answers,
and rationales; answers determined by geometry, visibility, or motion are
recomputed from the source records, and the resulting annotations are checked
against the paired images and motion logs. The benchmark contains 2,196 paired
frames, 35,136 question--answer records, and 2,129 complete-horizon planning
examples.

Leveraging \dataset{}, we build \model{}, a unified VLM baseline for
cooperative scene understanding and planning. Its language decoder answers
CDQA prompts and generates the
auxiliary rationale, while structured heads predict the planning command and
waypoints. Experiments show that CDQA adaptation improves question-answering
accuracy, while QA initialization and rationale supervision each reduce
planning error relative to training on planning targets alone. The
standard CP configuration of \model{} also achieves a lower final displacement
error than the compared V2X planners.

\paragraph{Contributions}
\begin{itemize}
\item We introduce \dataset{}, a real-world benchmark that aligns scene-grounded
  CDQA, auxiliary three-part rationales, and cooperative planning targets on
  synchronized vehicle--infrastructure scenes.
\item We establish \model{}, a unified VLM baseline that directly consumes
  paired vehicle--infrastructure images for cooperative understanding and
  structured ego-trajectory planning, reaching a lower FDE than the compared
  V2X planners.
\item We evaluate cooperative VLMs across task adaptation,
  supervision sources, backbone choices, and roadside-input conditions.
\end{itemize}

\section{Related Work}

\subsection{Driving QA and visual evidence}

Driving-VLM benchmarks extend autonomous-driving evaluation from geometry to
scene understanding, QA, reasoning, and planning
\citep{nuscenesqa,lingoqa,drivelm,reason2drive,wang2025omnidrive}. An answer
counts as understanding only if it depends on the images: plausible answers
persist under corrupted or missing vision
\citep{xie2025drivebench,liao2025robodrivevlm}, removing an object changes
what a planner does
\citep{panda2026cvaa,chahe2026whatshidden}, and
hallucination diagnostics measure the same gap outside driving
\citep{rohrbach2018chair,li2023pope}. This is why \dataset{} scores CDQA on
answers recomputed from the source records, and why both tasks are also
reported with the roadside input withheld or replaced.

Roadside and cooperative-driving datasets study grounding, QA, and reasoning
across infrastructure or multiple vehicle views
\citep{yang2025monirefer,guan2026roadscenevqa,zhou2025tumtrafficvideoqa,
chiu2025v2vllm,chiu2025v2vgot}. V2X-QA~\citep{you2026v2xqa} compares
vehicle-only, infrastructure-only, and cooperative views through
multiple-choice QA. D2-V2X~\citep{richard2026d2v2x} evaluates grounded
question--rationale--answer outputs for ego-occluded objects, while VGGDrive,
DriveSpatial, and CrossView emphasize cross-view geometry and spatiotemporal
reasoning \citep{wang2026vggdrive,vo2026drivespatial,shah2026crossview}.
\dataset{} carries planning
targets on the same paired scenes it annotates for CDQA, so what a model
answers and what it decides are measured on one set of frames.

\subsection{VLMs for driving and trajectory planning}

Driving VLMs connect scene understanding to commands or trajectories through
language-conditioned policies, discrete tokens, structured queries, and future
prediction \citep{vlp,drivegpt4,lmdrive,jiang2024senna,hwang2024emma,simlingo,
jiang2025alphadrive,fu2025orion,li2025recogdrive,zhou2025autovla,
li2025driver1,liu2026drivepi,yang2026drivemoe,tian2026ccot,
liu2026anchorvla,zheng2026drivema,zhang2026onedrive,wang2026vlaworld,
yu2026hyworldvla}. \model{} provides a unified VLM baseline for paired ego and
roadside images, with the same trajectory and command objectives across
backbone choices.

Some driving VLMs generate text and trajectories through separate outputs
\citep{simlingo,zhang2025reasoningvla,lu2026onevl}. Generated rationales need
not fully describe the model's internal decision process
\citep{turpin2023unfaithful,lanham2023faithfulness,
foutter2026faithfulness}. In \model{}, the rationale is an auxiliary training
target, and the structured heads predict commands and trajectories directly.

\subsection{V2X perception, planning, and cooperative evidence}

Cooperative perception shares raw measurements, detections, or learned
features to recover evidence outside a single agent's field of view
\citep{collaborative-survey2,Fcooper,OPV2V,v2vnet,where2comm,CoBEVT,
how2comm,v2x-vit}. Recent work improves selective, instance-aware, and
uncertainty-aware exchange alongside compressed communication and temporal
alignment
\citep{ACCO,EIMC,yang2026uecp,CodeFilling,CoBEVFlow,sparsealign,Traf-align}.
Heterogeneous collaboration further motivates extensible fusion and
preparation-free online adaptation \citep{HEAL,PolyInter,yang2026bolt}. These
methods primarily measure detection, bandwidth, or robustness. Paired
real-world observations come from DAIR-V2X, V2X-Seq, and V2X-Real
\citep{dair-v2x,v2x-seq,v2xreal}, and UniV2X, UniMM-V2X, MAP, and Defer to
Plan connect cooperative sensing to planning
\citep{yu2025univ2x,song2025unimmv2x,yin2025map,li2026defertoplan}.
MDrive~\citep{coscoy2026mdrive} evaluates closed-loop multi-agent behavior and
finds that stronger cooperative perception need not improve planning, which is
why \dataset{} scores understanding and decision on the same frames rather
than inferring one from the other.

Cooperative VLM and VLA research now spans paired-view planning,
language-based knowledge pools, latent infrastructure guidance, and
cooperative policy learning
\citep{you2024v2xvlm,you2025seal,liu2025colmdriver,peng2026omniv2x,
boroujeni2026vla4codrive,luo2026v2xunipool,song2026dhvlm}. V2X-QA controls
view availability for real-world multiple-choice QA~\citep{you2026v2xqa};
concurrent V2XBench, together with its AURORA planner, combines simulated V2X
VQA with closed-loop planning~\citep{xu2026roadsidecooperative}; CMU-Drive
with V2V-VLA studies closed-loop cooperative reasoning and policy learning
\citep{chiu2026cmudrive}; and Bench2Drive-VL plus nuReasoning broaden ego-view
evaluation toward closed-loop behavior and long-tail reasoning
\citep{jia2026bench2drivevl,huang2026nureasoning}. \dataset{} adds CDQA
annotations and three-part rationales to real-world vehicle-to-infrastructure
scenes that also carry CP targets.

Table~\ref{tab:benchmark-positioning} compares task and evaluation coverage
across these benchmarks; concurrent preprints are discussed above but not
tabulated.

\begin{table}[!htbp]
\caption{\textbf{Benchmark positioning}. V2X, Src, QA, CoT, and Plan denote
paired V2X input, source records, question answering, reasoning annotations, and
planning targets; our row evaluates CDQA and CP. Text denotes text-similarity metrics.
\Greenword{} checks mark explicit
benchmark support, \amberword{} half-circles mark a related signal without full joint
evaluation, and \redword{} crosses mark absent support.}
\label{tab:benchmark-positioning}
\centering
\small
\setlength{\tabcolsep}{4.8pt}
\renewcommand{\arraystretch}{1.08}
\begin{tabular}{lcccccl}
\toprule
Benchmark & V2X & Src & QA & CoT & Plan & Metrics \\
\midrule
DriveLM~\benchref{drivelm} & \xmark & \xmark & \cmark & \pmark & \pmark & Text/GPT \\
Reason2Drive~\benchref{reason2drive} & \xmark & \xmark & \cmark & \cmark & \pmark & Reasoning \\
DriveBench~\benchref{xie2025drivebench} & \xmark & \xmark & \cmark & \pmark & \pmark & Accuracy/Text/GPT \\
D2-V2X~\benchref{richard2026d2v2x} & \cmark & \cmark & \cmark & \cmark & \pmark & Grounding/Reasoning \\
V2X-Seq~\benchref{v2x-seq} & \cmark & \pmark & \xmark & \xmark & \pmark & Detection/Forecast \\
UniV2X~\benchref{yu2025univ2x} & \cmark & \pmark & \xmark & \xmark & \cmark & Planning \\
V2X-QA~\benchref{you2026v2xqa} & \cmark & \pmark & \cmark & \xmark & \pmark & MCQA/Calibration \\
CMU-Drive~\benchref{chiu2026cmudrive} & \cmark & \pmark & \xmark & \cmark & \cmark & Closed-loop/Reasoning \\
\midrule
\tablefocus
\textbf{\dataset{}} & \cmark & \cmark & \cmark & \cmark & \cmark & CDQA/CP \\
\bottomrule
\end{tabular}
\end{table}

\FloatBarrier

\section{\dataset{}: Benchmark Design}
\label{sec:benchmark}

\dataset{} is built on V2X-Seq-SPD~\citep{v2x-seq}, which supplies
time-synchronized ego and infrastructure images, cooperative object labels,
and ego-motion records. We organize 2,196 paired frames into the annotation
layers in Table~\ref{tab:annotation-layers}, with L3 providing CDQA supervision
and evaluation data. The 2,129 frames with complete three-second motion
records also carry CP targets and L4 rationales for auxiliary supervision.

\begin{table}[!htbp]
\caption{\textbf{Annotation layers in \dataset{}}.}
\label{tab:annotation-layers}
\centering
\small
\setlength{\tabcolsep}{4pt}
\renewcommand{\arraystretch}{1.08}
\begin{tabular}{@{}>{\raggedright\arraybackslash}p{0.27\linewidth}
>{\raggedright\arraybackslash}p{0.34\linewidth}
>{\raggedright\arraybackslash}p{\dimexpr0.39\linewidth-16pt\relax}@{}}
\toprule
Annotation layer & Content & Construction \\
\midrule
L1: Object evidence & Object attributes, source views, and motion states
& Label matching and track analysis \\
L2: Scene descriptions & Ego view, roadside view, and complementary observations
& Model drafting from paired images and L1 \\
L3: Question--answer annotations & Driving questions, answers, and reasons
& Template-based drafting and record-derived answers \\
L4: Three-part rationales & Scene overview, V2X-aware critical objects, and decision reasoning
& Model drafting with scene evidence \\
\bottomrule
\end{tabular}
\end{table}

\subsection{L1: Object evidence}

\begin{figure}[!t]
\centering
\includegraphics[width=\linewidth]{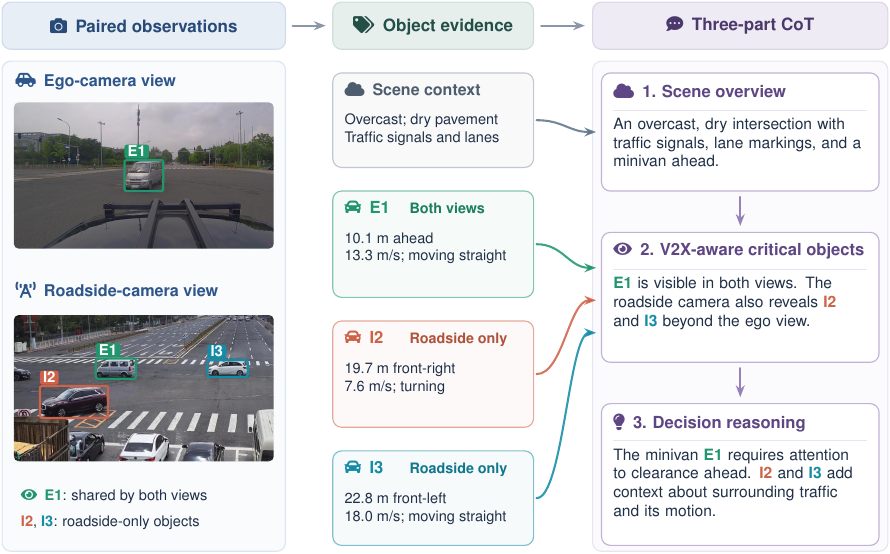}
\caption{\textbf{Source-traceable evidence chain}. Paired observations link to
object attributes and a three-part rationale summary. Matching colors identify
the shared object E1 and roadside-only objects I2 and I3 across views,
records, and text. The rationale separates the scene overview, the V2X-aware
critical objects, and the decision reasoning.}
\label{fig:v2x-example}
\end{figure}

The UniV2X SPD converter~\citep{yu2025univ2x} places cooperative boxes in the
ego frame. L1 assigns source views by matching these boxes to ego-side and
roadside labels within 3~m: both, ego-only, roadside-only, or unmatched when
no label falls within that threshold. Of 54,171 objects, 23,792 (43.9\%) are
matched only to the roadside labels, and such objects appear in 98.3\% of
frames (median 9; Appendix Figure~\ref{fig:evidence-statistics}). Tracks
around the frame provide motion states, distinguishing parking from temporary
stops that have the same instantaneous speed in the source records, as well as
turning, going-straight, and slow-moving states.

\subsection{L2--L3: Scene descriptions and question--answer annotations}

Qwen3-VL-32B drafts L2 descriptions from the paired
images and L1, covering the ego view, roadside view, and complementary roadside
observations. Using these descriptions, images, and object records, it drafts
L3 answers and reasons for 16 templates per frame across prediction, maneuver
QA, and V2X evidence analysis (35,136 records).
For eight templates, answers are computed from records by fixed rules rather
than drafted: the motion states give the two relative-distance answers, the
recorded ego trajectory gives the maneuver choice, and the annotated views and
the geometry give the roadside-only identities, distances, and criticality
answers. Five of them support the reported CDQA accuracy; the full set appears
in Appendix Table~\ref{tab:qa-annotation-catalog}.

\subsection{L4: Rationales}

Qwen3-VL-32B drafts three-part rationales from paired images and scene
evidence:
(1) a scene overview covering weather, road conditions, and traffic;
(2) V2X-aware critical objects with category, distance, bearing, motion, and
source view; and (3) decision reasoning that separates roadside contributions
from ego observations. Rationales serve as auxiliary supervision and generated
outputs rather than a scored task. Figure~\ref{fig:v2x-example} illustrates one
example; Appendix~\ref{sec:appendix-annotation-validation} details generation inputs.

\paragraph{Planning targets.}
We transform subsequent recorded ego positions into the current ego coordinate
frame and sample six waypoints at 0.5-second intervals through 3.0 seconds.
The full-horizon requirement yields 2,129 CP examples. Fixed rules derive a
six-class command from the motion (Appendix~\ref{sec:appendix-metric-notes}).

\paragraph{Annotation checks.}
Every scene description, question--answer record, and rationale was
human-verified against the paired images and source records. Rejected items
were corrected or regenerated and inspected again;
Appendix~\ref{sec:appendix-annotation-validation} details the checks.

\FloatBarrier
\section{\model{}: A Unified VLM Baseline}
\label{sec:method}

\begin{figure}[!htbp]
    \centering
    \includegraphics[width=0.96\linewidth]{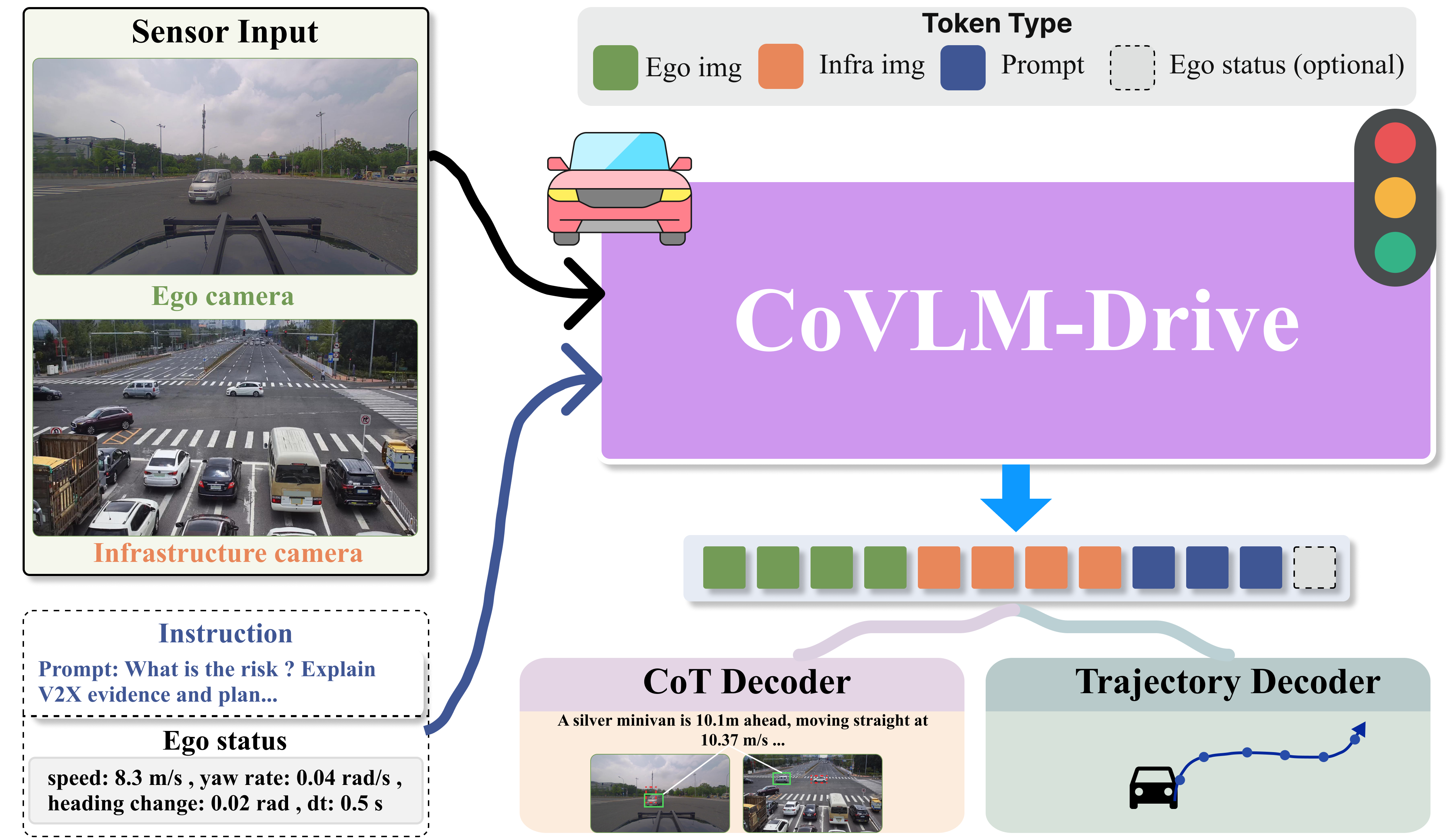}
\caption{Planning path of \model{}. The ego image, roadside image, and prompt
enter one VLM backbone. The language decoder generates the three-part
rationale; two structured heads predict the driving command and six
ego-relative waypoints. The same language decoder supports CDQA responses.
The dashed ego-status input is used by the optional variant marked with an
asterisk in Table~\ref{tab:external-baselines}.}
\label{fig:framework}
\end{figure}

\model{} uses a single VLM backbone to process the ego image
\(\mathbf{I}^{e}\), roadside image \(\mathbf{I}^{r}\), and task prompt \(q\)
for CDQA and CP. Its language decoder generates CDQA answers or three-part
rationales, while two structured heads predict the command class \(c\) and
ego-relative waypoint sequence \(\mathbf{P}\).
Figure~\ref{fig:framework} illustrates the planning path.

\subsection{Paired-view encoding and task outputs}

We place the ego image, roadside image, and prompt in a
single Qwen3-VL input sequence and adapt the backbone with LoRA
\citep{bai2025qwen3vl,hu2022lora}.

Let \(\mathbf{h}_{1:L}\) denote the final hidden states of a sequence of
length \(L\). The mask \(m_l\) equals one for prompt tokens, including images,
and zero for target tokens and padding. The structured heads use their
masked mean:
\begin{equation}
\bar{\mathbf{h}}=
\frac{\sum_{l=1}^{L}m_l\mathbf{h}_l}{\sum_{l=1}^{L}m_l}.
\end{equation}
The waypoint MLP maps \(\bar{\mathbf{h}}\) to six ego-relative planar
waypoints, and the command MLP maps it to a command distribution.

\subsection{Task supervision and inference}
\label{sec:training-objective-interface}

For CDQA, we adapt the backbone with LoRA using autoregressive cross-entropy
on the benchmark's question--answer annotations. For CP, we
jointly train the backbone adapters and structured heads with rationale,
command, and trajectory supervision:
\begin{equation}
\mathcal{L} =
\lambda_{\mathrm{lm}}\mathcal{L}_{\mathrm{lm}} +
\lambda_{\mathrm{cmd}}\mathcal{L}_{\mathrm{cmd}} +
\lambda_{\mathrm{wp}}\mathcal{L}_{\mathrm{wp}} +
\lambda_{\mathrm{fde}}\mathcal{L}_{\mathrm{fde}},
\end{equation}
where \(\mathcal{L}\) is the total loss, the \(\lambda\) terms are scalar
weights, \(\mathcal{L}_{\mathrm{lm}}\) supervises the generated rationale,
\(\mathcal{L}_{\mathrm{cmd}}\) supervises command classification, and
\(\mathcal{L}_{\mathrm{wp}}\) and \(\mathcal{L}_{\mathrm{fde}}\) supervise
the waypoint sequence and its endpoint. The rationale loss is token-level
cross-entropy. Command classification uses cross-entropy, and waypoint
regression uses Smooth L1 with an additional endpoint term. Rationale text is
supervised through teacher forcing, while the planning heads use the mean
hidden representation of the image and prompt tokens. Optionally, CP training
can initialize the backbone adapters from CDQA fine-tuning;
Section~\ref{sec:supervision-effects} evaluates this training choice, with
full metrics in Appendix~\ref{sec:appendix-supervision-ablations}.

At inference, the language decoder generates answers to CDQA prompts. For CP,
the structured heads directly map the pooled prompt representation to a
command and trajectory. The language decoder can also generate a three-part
rationale from the paired observations.

\FloatBarrier

\section{Experiments}

\subsection{Experimental setup}
\label{sec:experimental-setup}

We evaluate CDQA and CP on the benchmark's scene-disjoint split.

All planning rows use the same 654 examples and ego-relative waypoint
convention. Table~\ref{tab:external-baselines} compares trained V2X planners,
\model{} with different backbones, and zero-shot VLMs prompted to generate
commands and waypoints. UniV2X uses released predictions; UniMM-V2X and MAP
use protocol-aligned reproductions. CDQA compares eight zero-shot VLMs and
two adapted \model{} checkpoints based on Qwen3-VL-8B: CP-adapted uses
rationale and planning supervision, and CDQA-adapted uses supervised
fine-tuning (SFT) on the question--answer annotations.

At horizon \(t\), L2 is the mean Euclidean distance between predicted and
reference waypoints; final displacement error (FDE) is L2 at \(3.0\)~s. Command
accuracy (Cmd Acc.) measures command classification, and balanced accuracy
(Bal. Acc.) averages recall over the command classes present in the validation
set. CDQA reports exact-answer accuracy over five objective tasks on the same
2,861 eligible items per input condition for every model. Maneuver QA compares
answers with the recorded ego maneuver, derived from the future trajectory by
a fixed rule. Object identifiers are neutral with respect to camera source.
Efficiency measurements use batch size one on an RTX~4090 and report model
computation time. Appendix~\ref{sec:appendix-reproducibility-release} gives
training, runtime, scoring, and baseline-alignment details;
Appendix~\ref{sec:appendix-communication} analyzes transmission cost.

\subsection{CDQA results}

\begin{table}[!t]
\caption{\textbf{CDQA results} on 2,861 eligible items per condition from five
objective tasks. The two \model{} checkpoints use CP supervision or CDQA SFT,
respectively. Best absent is the better of ego-only and blank (named in
parentheses); Shuffled uses a mismatched roadside scene.
$\Delta$ is Valid minus Best absent; \greenword{}/\redword{} mark gaps of
one point or more.}
\label{tab:main-l3qa}
\centering
\begingroup
\small
\setlength{\tabcolsep}{6.6pt}
\renewcommand{\arraystretch}{1.08}
\begin{tabular}{lcccc}
\toprule
Model & Valid (\%) & Best absent (\%) & Shuffled (\%) & $\Delta$ (pp) \\
\midrule
\tablegroupgray{5}{\benchicon{benchGray}{globe}\enspace Zero-shot VLMs}
Qwen3-VL-4B & 20.4 & \qactrl{14.2}{Ego} & 13.7 & \qagood{+6.1} \\
Qwen3-VL-8B & 24.7 & \qactrl{19.5}{Ego} & 19.6 & \qagood{+5.2} \\
Qwen3-VL-32B & 51.1 & \qactrl{17.8}{Blk} & 17.5 & \qagood{+33.2} \\
Qwen2.5-VL-3B & 24.9 & \qactrl{19.3}{Ego} & 22.4 & \qagood{+5.6} \\
Qwen2.5-VL-7B & 19.4 & \qactrl{12.5}{Ego} & 13.1 & \qagood{+6.9} \\
InternVL3-8B & 29.8 & \qactrl{19.0}{Ego} & 19.2 & \qagood{+10.9} \\
LLaVA-OneVision-7B & 23.4 & \qactrl{24.4}{Blk} & 21.6 & \qabad{-1.0} \\
Idefics3-8B & 16.9 & \qactrl{9.8}{Ego} & 4.1 & \qagood{+7.1} \\
\tablegroupblue{5}{\benchicon{benchBlue}{route}\enspace Adapted Qwen3-VL-8B models}
CoVLM-Drive (CP-adapted) & 25.8 & \qactrl{20.8}{Ego} & 21.1 & \qagood{+5.1} \\
CoVLM-Drive (CDQA-adapted) & 39.6 & \qactrl{14.2}{Ego} & 15.7 & \qagood{+25.3} \\
\bottomrule
\end{tabular}
\endgroup

\end{table}

Table~\ref{tab:main-l3qa} compares eight zero-shot VLMs and the two adapted
\model{} checkpoints. With matched paired views, zero-shot accuracy ranges
from 16.9\% to 51.1\%; Qwen3-VL-32B, also used for annotation drafting,
achieves the highest accuracy. CDQA adaptation raises the Qwen3-VL-8B
backbone from 24.7\% to 39.6\%, a gain of 14.9 percentage points. The
CP-adapted checkpoint scores 25.8\% after rationale and planning training.
These comparisons establish both a zero-shot reference and the training value
of the benchmark's CDQA annotations. Appendix~\ref{sec:appendix-l3qa-diagnostics}
provides task-grouped results and qualitative examples.

\subsection{CP results and backbone comparison}

Table~\ref{tab:external-baselines} compares CP accuracy and inference
cost. Backbone variants retain the same LoRA configuration, structured heads,
and training objectives; the starred variant additionally uses ego status.

\begin{table}[!t]
\caption{\textbf{CP results} on 654 examples. All groups share the
same trajectory and command metrics. Parse is the percentage of zero-shot
responses containing a valid command and six waypoints. Efficiency measures
structured planning inference; dashes denote unreported or inapplicable
entries. Bold/underline mark best/second-best trained results. \(\dagger\)
denotes our reproduction; \blueword{} marks the standard CP configuration and
\grayword{} its ego-status variant (\(\ast\)).}
\label{tab:external-baselines}
\centering
\scriptsize
\setlength{\tabcolsep}{1.55pt}
\renewcommand{\arraystretch}{0.94}
\begin{adjustbox}{max width=\linewidth}
\begin{tabular}{lrrrrrrrrrrrr}
\toprule
& \multicolumn{6}{c}{Trajectory L2 (m) $\downarrow$}
& \multicolumn{2}{c}{Command $\uparrow$}
& \multicolumn{1}{c}{Output}
& \multicolumn{3}{c}{Efficiency} \\
\cmidrule(lr){2-7}\cmidrule(lr){8-9}\cmidrule(lr){10-10}\cmidrule(l){11-13}
Method & 0.5 s & 1.0 s & 1.5 s & 2.0 s & 2.5 s & FDE
& Acc. & Bal. & \makecell{Parse\\(\%)} & \makecell{Lat. (ms)\\$\downarrow$} & FPS $\uparrow$
& \makecell{Mem. (GiB)\\$\downarrow$} \\
\midrule
\tablegroupgray{13}{Task-trained V2X planners}
UniV2X~\benchref{yu2025univ2x}
& 1.245 & 2.079 & 2.888 & 3.831 & 4.843 & 5.861
& -- & -- & -- & 812.7 & 1.23 & \runnerup{2.86} \\
UniMM-V2X$^\dagger$~\benchref{song2025unimmv2x}
& 0.893 & 1.677 & 2.528 & 3.427 & 4.336 & 5.357
& \runnerup{0.8165} & \runnerup{0.6604} & -- & 523.9
& 1.91 & 3.29 \\
MAP$^\dagger$~\benchref{yin2025map}
& \textbf{0.601} & \textbf{1.408} & \runnerup{2.297}
& 3.233 & 4.200 & 5.240
& \textbf{0.8394} & \textbf{0.7297} & -- & 588.6 & 1.70 & \textbf{2.71} \\
\addlinespace[1pt]
\tablegroupblue{13}{\model{} backbone variants}
\tablefocus
\textbf{Qwen3-VL-8B (standard CP)}
& 0.812 & 1.500 & \textbf{2.283}
& \textbf{3.119} & \textbf{3.998} & \textbf{4.911}
& 0.7706 & 0.5661 & -- & \runnerup{223.1} & \runnerup{4.48} & 17.08 \\
\rowcolor[gray]{0.88}
Qwen3-VL-8B + ego status$^\ast$
& \runnerup{0.798} & 1.515 & 2.330 & 3.207 & 4.136 & 5.126
& 0.7401 & 0.5497 & -- & \textbf{180.1} & \textbf{5.55} & 17.11 \\
Qwen2.5-VL-7B
& 0.832 & 1.530 & 2.364 & 3.241 & 4.165 & 5.120
& 0.7569 & 0.5355 & -- & 273.8 & 3.65 & 16.23 \\
InternVL3-8B
& 0.811 & \runnerup{1.493} & 2.310 & \runnerup{3.145}
& \runnerup{4.077} & \runnerup{5.076}
& 0.7324 & 0.5557 & -- & 305.1 & 3.28 & 16.08 \\
LLaVA-OneVision-7B
& 0.825 & 1.555 & 2.388 & 3.284 & 4.238 & 5.272
& 0.7385 & 0.5309 & -- & 280.9 & 3.56 & 16.19 \\
Idefics3-8B
& 0.876 & 1.691 & 2.584 & 3.525 & 4.519 & 5.538
& 0.7492 & 0.5850 & -- & 1070.6 & 0.93 & 19.24 \\
\addlinespace[1pt]
\tablegroupgray{13}{Zero-shot VLMs}
Qwen3-VL-4B
& 3.853 & 7.251 & 10.986 & 14.827 & 18.754 & 22.751
& 0.6055 & 0.2235 & 96.3 & -- & -- & -- \\
Qwen3-VL-8B
& 1.911 & 3.640 & 5.444 & 7.311 & 9.240 & 11.227
& 0.6468 & 0.2387 & 100.0 & -- & -- & -- \\
Qwen2.5-VL-3B
& 3.996 & 6.784 & 10.172 & 13.582 & 17.000 & 20.426
& 0.2202 & 0.0813 & 34.4 & -- & -- & -- \\
Qwen2.5-VL-7B
& 3.803 & 7.132 & 10.618 & 14.153 & 17.711 & 21.284
& 0.6774 & 0.2500 & 100.0 & -- & -- & -- \\
InternVL3-8B
& 3.581 & 6.828 & 10.108 & 13.397 & 16.698 & 20.007
& 0.6774 & 0.2500 & 100.0 & -- & -- & -- \\
LLaVA-OneVision-7B
& 5.324 & 77.449 & 152.437 & 227.498 & 302.597 & 377.721
& 0.6774 & 0.2500 & 100.0 & -- & -- & -- \\
Idefics3-8B
& 9.099 & 10.854 & 14.049 & 17.434 & 21.606 & 25.945
& 0.6606 & 0.2438 & 99.8 & -- & -- & -- \\
\bottomrule
\end{tabular}
\end{adjustbox}
\end{table}

The standard CP configuration achieves the lowest trajectory error from 1.5~s through
the endpoint, with 4.911~m FDE compared with 5.240~m for MAP and 5.357~m for
UniMM-V2X.

The backbone comparison establishes reference results for the same planning
framework across VLM families. Alternative backbones achieve 5.076--5.538~m
FDE, while the best zero-shot result is 11.227~m. The standard CP
configuration runs at 223.1~ms per frame and 4.48~FPS, more than twice the
throughput of the compared V2X planners.

\subsection{Effect of QA and rationale supervision}
\label{sec:supervision-effects}

Table~\ref{tab:supervision-summary} compares QA initialization and rationale
supervision for \model{}. Relative to planning targets alone (5.197~m FDE), QA
initialization reduces FDE to 5.083~m and rationale supervision reduces it to
4.911~m; QA initialization also raises CDQA accuracy after CP training. The
standard CP configuration uses rationale supervision alone.
Appendix~\ref{sec:appendix-supervision-ablations} reports the full factorial
with per-horizon and command metrics.

\begin{table}[!htbp]
\caption{\textbf{QA initialization and rationale supervision}. All rows use
Qwen3-VL-8B, paired images, the same planning heads and CP training budget,
and no ego-status input. QA initialization adds a prior CDQA adaptation stage.
FDE uses 654 CP examples; CDQA accuracy is measured after CP training on
2,861 items. Bold marks the best result in each column; \blueword{} marks
the standard CP configuration.}
\label{tab:supervision-summary}
\centering
\small
\setlength{\tabcolsep}{8pt}
\renewcommand{\arraystretch}{1.08}
\begin{tabular}{ccrr}
\toprule
QA initialization & Rationale supervision & FDE (m) $\downarrow$ & CDQA accuracy (\%) $\uparrow$ \\
\midrule
No & No & 5.197 & 24.8 \\
\tablefocus
No & Yes & \textbf{4.911} & 25.8 \\
Yes & No & 5.083 & \textbf{37.4} \\
\bottomrule
\end{tabular}
\end{table}

\subsection{Effect of roadside observations}
\label{sec:roadside-input-planning}

We hold the sample, ego image, target, and model fixed while varying the
roadside input: ego-only removes the view, blank replaces it with an empty
image, and shuffled supplies a different scene. Appendix
Figure~\ref{fig:matched-controls} illustrates these conditions.

For CDQA (Table~\ref{tab:main-l3qa}), matched input outperforms the best
absent-view control for seven of eight zero-shot VLMs and outperforms shuffled
input for all ten models. CDQA adaptation increases the gap over the best
absent-view control from 5.2 to 25.3 percentage points.
Appendix~\ref{sec:appendix-l3qa-diagnostics} breaks down this effect by task.

For CP (Table~\ref{tab:input-control-planning}), the standard CP configuration
obtains 4.911~m FDE with matched input, compared with 5.776~m for ego-only,
5.085~m for blank, and 4.976~m for shuffled input. The matched view gives the
lowest endpoint error, with the largest difference occurring when the roadside
view is removed. Shuffling the roadside view moves the predicted waypoints by
a mean of 0.73~m, exceeding 0.1~m on 95.3\% of examples, so the planner
conditions on roadside content. Appendix~\ref{sec:appendix-planning-controls}
provides the corresponding analysis on the subset whose rationales flag a
roadside-only critical object.

\begin{table}[!htbp]
\caption{\textbf{Roadside-input controls} on the same 654 examples. FDE is the
endpoint at 3.0~s and $\Delta$FDE is control minus valid, computed before
rounding; Acc. and Bal. are command accuracy and balanced accuracy.}
\label{tab:input-control-planning}
\centering
\small
\setlength{\tabcolsep}{4.2pt}
\renewcommand{\arraystretch}{1.04}
\begin{adjustbox}{max width=\linewidth}
\begin{tabular}{lrrrrrrrrr}
\toprule
& \multicolumn{7}{c}{Trajectory L2 (m) $\downarrow$}
& \multicolumn{2}{c}{Command $\uparrow$} \\
\cmidrule(lr){2-8}\cmidrule(l){9-10}
Roadside input & 0.5 s & 1.0 s & 1.5 s & 2.0 s & 2.5 s & FDE
& $\Delta$FDE & Acc. & Bal. \\
\midrule
\tablefocus
Valid & 0.812 & 1.500 & 2.283 & 3.119 & 3.998
& \bestcell{4.911} & Ref. & 0.7706 & 0.5661 \\
\midrule
Ego-only & 0.837 & 1.622 & 2.529 & 3.606 & 4.693
& 5.776 & +0.864 & 0.7141 & 0.3679 \\
Blank & 0.795 & 1.467 & 2.321 & 3.224 & 4.128
& 5.085 & +0.174 & 0.7875 & 0.5567 \\
Shuffled & 0.822 & 1.504 & 2.298 & 3.158 & 4.047
& 4.976 & +0.065 & 0.7645 & 0.5340 \\
\bottomrule
\end{tabular}
\end{adjustbox}
\end{table}

\FloatBarrier
\section{Conclusion}

We introduced \dataset{}, which connects scene-grounded CDQA annotations and
ego-trajectory targets on real-world vehicle--infrastructure observations, and
\model{}, a unified VLM baseline for both tasks. CDQA adaptation improves
answer accuracy, while the standard CP configuration achieves a lower FDE than
the compared V2X planners. QA initialization and rationale supervision each
reduce planning error. Together, the benchmark and baseline support learning
and evaluating cooperative scene understanding and planning.

\clearpage

\bibliography{iclr2027_conference}

\begin{thebibliography}{86}
\providecommand{\natexlab}[1]{#1}
\providecommand{\url}[1]{\texttt{#1}}
\expandafter\ifx\csname urlstyle\endcsname\relax
  \providecommand{\doi}[1]{doi: #1}\else
  \providecommand{\doi}{doi: \begingroup \urlstyle{rm}\Url}\fi

\bibitem[Bai et~al.(2025)Bai, Cai, Chen, Chen, Chen, Cheng, Deng, Ding, Gao,
  Ge, Ge, Guo, Huang, Huang, Huang, Hui, Jiang, Li, Li, Li, Li, Lin, Lin, Liu,
  Liu, Liu, Liu, Liu, Liu, Lu, Luo, Lv, Men, Meng, Ren, Ren, Song, Sun, Tang,
  Tu, Wan, Wang, Wang, Wang, Wang, Xie, Xu, Xu, Xu, Yang, Yang, Yang, Yang, Yu,
  Zhang, Zhang, Zhang, Zheng, Zhong, Zhou, Zhou, Zhou, Zhu, and
  Zhu]{bai2025qwen3vl}
Shuai Bai, Yuxuan Cai, Ruizhe Chen, Keqin Chen, Xionghui Chen, Zesen Cheng,
  Lianghao Deng, Wei Ding, Chang Gao, Chunjiang Ge, Wenbin Ge, Zhifang Guo,
  Qidong Huang, Jie Huang, Fei Huang, Binyuan Hui, Shutong Jiang, Zhaohai Li,
  Mingsheng Li, Mei Li, Kaixin Li, Zicheng Lin, Junyang Lin, Xuejing Liu,
  Jiawei Liu, Chenglong Liu, Yang Liu, Dayiheng Liu, Shixuan Liu, Dunjie Lu,
  Ruilin Luo, Chenxu Lv, Rui Men, Lingchen Meng, Xuancheng Ren, Xingzhang Ren,
  Sibo Song, Yuchong Sun, Jun Tang, Jianhong Tu, Jianqiang Wan, Peng Wang,
  Pengfei Wang, Qiuyue Wang, Yuxuan Wang, Tianbao Xie, Yiheng Xu, Haiyang Xu,
  Jin Xu, Zhibo Yang, Mingkun Yang, Jianxin Yang, An~Yang, Bowen Yu, Fei Zhang,
  Hang Zhang, Xi~Zhang, Bo~Zheng, Humen Zhong, Jingren Zhou, Fan Zhou, Jing
  Zhou, Yuanzhi Zhu, and Ke~Zhu.
\newblock {Qwen3-VL} technical report.
\newblock \emph{arXiv preprint arXiv:2511.21631}, 2025.
\newblock \doi{10.48550/arXiv.2511.21631}.
\newblock URL \url{https://arxiv.org/abs/2511.21631}.

\bibitem[Boroujeni \& Razi(2026)Boroujeni and Razi]{boroujeni2026vla4codrive}
Sayed Pedram~Haeri Boroujeni and Abolfazl Razi.
\newblock {VLA4CoDrive}: Vision-language-action dataset for cooperative
  autonomous driving.
\newblock In \emph{Proceedings of the IEEE/CVF Winter Conference on
  Applications of Computer Vision (WACV) Workshops}, pp.\  1789--1799, March
  2026.
\newblock URL
  \url{https://openaccess.thecvf.com/content/WACV2026W/LLVM-AD/html/Boroujeni_VLA4CoDrive_Vision-Language-Action_Dataset_for_Cooperative_Autonomous_Driving_WACVW_2026_paper.html}.

\bibitem[Chahe et~al.(2026)Chahe, Naes, D'sa, Tariq, Bae, Zhou, and
  Isele]{chahe2026whatshidden}
Amirhosein Chahe, Tyler Naes, Jovin D'sa, Faizan~M. Tariq, Sangjae Bae, Lifeng
  Zhou, and David Isele.
\newblock What's hidden matters: Identifying planning-critical occluded agents
  using vision-language models.
\newblock In \emph{IEEE/RSJ International Conference on Intelligent Robots and
  Systems}, 2026.
\newblock URL \url{https://arxiv.org/abs/2607.00283}.

\bibitem[Chen et~al.(2019)Chen, Ma, Tang, Guo, Yang, and Fu]{Fcooper}
Qi~Chen, Xu~Ma, Sihai Tang, Jingda Guo, Qing Yang, and Song Fu.
\newblock {F-cooper}: Feature based cooperative perception for autonomous
  vehicle edge computing system using {3D} point clouds.
\newblock In \emph{Proceedings of the 4th ACM/IEEE Symposium on Edge
  Computing}, pp.\  88--100, 2019.
\newblock \doi{10.1145/3318216.3363300}.

\bibitem[Chiu \& Smith(2026)Chiu and Smith]{chiu2026cmudrive}
Hsu-kuang Chiu and Stephen~F. Smith.
\newblock {CMU-Drive and V2V-VLA}: Cooperative multi-agent unified driving with
  reasoning benchmark and vehicle-to-vehicle vision-language-action models,
  2026.
\newblock URL \url{https://arxiv.org/abs/2608.07621}.

\bibitem[Chiu et~al.(2025{\natexlab{a}})Chiu, Hachiuma, Wang, Smith, Wang, and
  Chen]{chiu2025v2vllm}
Hsu-kuang Chiu, Ryo Hachiuma, Chien-Yi Wang, Stephen~F. Smith, Yu-Chiang~Frank
  Wang, and Min-Hung Chen.
\newblock {V2V-LLM}: Vehicle-to-vehicle cooperative autonomous driving with
  multimodal large language models, 2025{\natexlab{a}}.
\newblock URL \url{https://arxiv.org/abs/2502.09980}.

\bibitem[Chiu et~al.(2025{\natexlab{b}})Chiu, Hachiuma, Wang, Wang, Chen, and
  Smith]{chiu2025v2vgot}
Hsu-kuang Chiu, Ryo Hachiuma, Chien-Yi Wang, Yu-Chiang~Frank Wang, Min-Hung
  Chen, and Stephen~F. Smith.
\newblock {V2V-GoT}: Vehicle-to-vehicle cooperative autonomous driving with
  multimodal large language models and graph-of-thoughts, 2025{\natexlab{b}}.
\newblock URL \url{https://arxiv.org/abs/2509.18053}.

\bibitem[Coscoy et~al.(2026)Coscoy, Zhou, Zhao, Wei, Magtoto, Liu, Song,
  Zimmer, Huang, Tang, Zhou, and Ma]{coscoy2026mdrive}
Marco Coscoy, Zewei Zhou, Seth~Z. Zhao, Henry Wei, Angela Magtoto, Johnson Liu,
  Rui Song, Walter Zimmer, Zhiyu Huang, Chen Tang, Bolei Zhou, and Jiaqi Ma.
\newblock {MDrive}: Benchmarking closed-loop cooperative driving for end-to-end
  multi-agent systems, 2026.
\newblock URL \url{https://arxiv.org/abs/2605.10904}.

\bibitem[Fezeu et~al.(2023)Fezeu, Carpenter, Fiandrino, Ramadan, Ye, Widmer,
  Qian, and Zhang]{fezeu2023midband}
Rostand A.~K. Fezeu, Jason Carpenter, Claudio Fiandrino, Eman Ramadan, Wei Ye,
  Joerg Widmer, Feng Qian, and Zhi-Li Zhang.
\newblock Mid-band {5G}: A measurement study in europe and {US}.
\newblock \emph{arXiv preprint arXiv:2310.11000}, 2023.
\newblock URL \url{https://arxiv.org/abs/2310.11000}.

\bibitem[Foutter et~al.(2026)Foutter, Cercola, Wild, Wang, Li, Gammelli, and
  Pavone]{foutter2026faithfulness}
Matthew Foutter, Matteo Cercola, Lena Wild, Yunshan Wang, Michelle Li, Daniele
  Gammelli, and Marco Pavone.
\newblock Do vision-language-action models mean what they say? on the role of
  faithfulness in embodied reasoning, 2026.
\newblock URL \url{https://arxiv.org/abs/2607.04681}.

\bibitem[Fu et~al.(2025)Fu, Zhang, Zhao, Cui, Liang, Zhang, Zhang, Xie, Wang,
  and Bai]{fu2025orion}
Haoyu Fu, Diankun Zhang, Zongchuang Zhao, Jianfeng Cui, Dingkang Liang, Chong
  Zhang, Dingyuan Zhang, Hongwei Xie, Bing Wang, and Xiang Bai.
\newblock {ORION}: A holistic end-to-end autonomous driving framework by
  vision-language instructed action generation.
\newblock In \emph{Proceedings of the IEEE/CVF International Conference on
  Computer Vision}, pp.\  24823--24834, October 2025.
\newblock \doi{10.1109/ICCV51701.2025.02302}.
\newblock URL
  \url{https://openaccess.thecvf.com/content/ICCV2025/html/Fu_ORION_A_Holistic_End-to-End_Autonomous_Driving_Framework_by_Vision-Language_Instructed_ICCV_2025_paper.html}.

\bibitem[Guan et~al.(2026)Guan, Hu, Chen, Xiao, Xia, Liu, Chen, Tang, Ouyang,
  Liang, Fan, Sun, and Yue]{guan2026roadscenevqa}
Runwei Guan, Rongsheng Hu, Shangshu Chen, Ningyuan Xiao, Xue Xia, Jiayang Liu,
  Beibei Chen, Ziren Tang, Ningwei Ouyang, Shaofeng Liang, Yuxuan Fan, Wanjie
  Sun, and Yutao Yue.
\newblock {RoadSceneVQA}: Benchmarking visual question answering in roadside
  perception systems for intelligent transportation system.
\newblock In \emph{Proceedings of the AAAI Conference on Artificial
  Intelligence}, volume~40, pp.\  4366--4375, 2026.
\newblock \doi{10.1609/aaai.v40i6.42434}.
\newblock URL \url{https://ojs.aaai.org/index.php/AAAI/article/view/42434}.

\bibitem[Han et~al.(2023)Han, Zhang, Li, Jin, Lang, and
  Li]{collaborative-survey2}
Yushan Han, Hui Zhang, Huifang Li, Yi~Jin, Congyan Lang, and Yidong Li.
\newblock Collaborative perception in autonomous driving: Methods, datasets,
  and challenges.
\newblock \emph{IEEE Intelligent Transportation Systems Magazine}, 15\penalty0
  (6):\penalty0 131--151, 2023.
\newblock \doi{10.1109/MITS.2023.3298534}.

\bibitem[Hu et~al.(2022{\natexlab{a}})Hu, Shen, Wallis, Allen-Zhu, Li, Wang,
  Wang, and Chen]{hu2022lora}
Edward~J. Hu, Yelong Shen, Phillip Wallis, Zeyuan Allen-Zhu, Yuanzhi Li, Shean
  Wang, Lu~Wang, and Weizhu Chen.
\newblock {LoRA}: Low-rank adaptation of large language models.
\newblock In \emph{ICLR}, 2022{\natexlab{a}}.

\bibitem[Hu et~al.(2022{\natexlab{b}})Hu, Fang, Lei, Zhong, and
  Chen]{where2comm}
Yue Hu, Shaoheng Fang, Zixing Lei, Yiqi Zhong, and Siheng Chen.
\newblock {Where2comm}: Communication-efficient collaborative perception via
  spatial confidence maps.
\newblock In \emph{Advances in Neural Information Processing Systems},
  volume~35, pp.\  4874--4886, 2022{\natexlab{b}}.

\bibitem[Hu et~al.(2024)Hu, Peng, Liu, Ge, Liu, and Chen]{CodeFilling}
Yue Hu, Juntong Peng, Sifei Liu, Junhao Ge, Si~Liu, and Siheng Chen.
\newblock Communication-efficient collaborative perception via information
  filling with codebook.
\newblock In \emph{Proceedings of the IEEE/CVF Conference on Computer Vision
  and Pattern Recognition}, pp.\  15481--15490, 2024.
\newblock \doi{10.1109/CVPR52733.2024.01466}.

\bibitem[Huang et~al.(2026)Huang, Liu, Song, Zhou, Yang, Zhang, Cai, Zhang,
  Gao, Xu, Chen, Shen, Guo, Qi, and Ma]{huang2026nureasoning}
Zhiyu Huang, Johnson Liu, Rui Song, Zewei Zhou, Ruining Yang, Yun Zhang,
  Tianhui Cai, Hanyin Zhang, Mingxuan Gao, Valeria Xu, Jiali Chen, Yishan Shen,
  Yiluan Guo, Tony~(Xuewei) Qi, and Jiaqi Ma.
\newblock {nuReasoning}: A reasoning-centric dataset and benchmark for
  long-tail autonomous driving, 2026.
\newblock URL \url{https://arxiv.org/abs/2605.31572}.

\bibitem[Hwang et~al.(2024)Hwang, Xu, Lin, Hung, Ji, Choi, Huang, He,
  Covington, Sapp, Zhou, Guo, Anguelov, and Tan]{hwang2024emma}
Jyh-Jing Hwang, Runsheng Xu, Hubert Lin, Wei-Chih Hung, Jingwei Ji, Kristy
  Choi, Di~Huang, Tong He, Paul Covington, Benjamin Sapp, Yin Zhou, James Guo,
  Dragomir Anguelov, and Mingxing Tan.
\newblock {EMMA}: End-to-end multimodal model for autonomous driving.
\newblock \emph{arXiv preprint arXiv:2410.23262}, 2024.

\bibitem[Jia et~al.(2026)Jia, Shao, Yang, Li, Zhang, and
  Yan]{jia2026bench2drivevl}
Xiaosong Jia, Yuqian Shao, Zhenjie Yang, Qifeng Li, Zhiyuan Zhang, and Junchi
  Yan.
\newblock {Bench2Drive-VL}: Benchmarks for closed-loop autonomous driving with
  vision-language models, 2026.
\newblock URL \url{https://arxiv.org/abs/2604.01259}.

\bibitem[Jiang et~al.(2024)Jiang, Chen, Liao, Zhang, Yin, Zhang, Huang, Liu,
  and Wang]{jiang2024senna}
Bo~Jiang, Shaoyu Chen, Bencheng Liao, Xingyu Zhang, Wei Yin, Qian Zhang, Chang
  Huang, Wenyu Liu, and Xinggang Wang.
\newblock {Senna}: Bridging large vision-language models and end-to-end
  autonomous driving.
\newblock \emph{arXiv preprint arXiv:2410.22313}, 2024.

\bibitem[Jiang et~al.(2025)Jiang, Chen, Zhang, Liu, and
  Wang]{jiang2025alphadrive}
Bo~Jiang, Shaoyu Chen, Qian Zhang, Wenyu Liu, and Xinggang Wang.
\newblock {AlphaDrive}: Unleashing the power of {VLMs} in autonomous driving
  via reinforcement learning and reasoning.
\newblock \emph{arXiv preprint arXiv:2503.07608}, 2025.
\newblock URL \url{https://arxiv.org/abs/2503.07608}.

\bibitem[Lanham et~al.(2023)Lanham, Chen, Radhakrishnan, Steiner, Denison,
  Hernandez, Li, Durmus, Hubinger, Kernion, Luko{\v{s}}i{\={u}}t{\.{e}},
  Nguyen, Cheng, Joseph, Schiefer, Rausch, Larson, McCandlish, Kundu, Kadavath,
  Yang, Henighan, Maxwell, Telleen-Lawton, Hume, Hatfield-Dodds, Kaplan,
  Brauner, Bowman, and Perez]{lanham2023faithfulness}
Tamera Lanham, Anna Chen, Ansh Radhakrishnan, Benoit Steiner, Carson Denison,
  Danny Hernandez, Dustin Li, Esin Durmus, Evan Hubinger, Jackson Kernion,
  Kamil{\.{e}} Luko{\v{s}}i{\={u}}t{\.{e}}, Karina Nguyen, Newton Cheng,
  Nicholas Joseph, Nicholas Schiefer, Oliver Rausch, Robin Larson, Sam
  McCandlish, Sandipan Kundu, Saurav Kadavath, Shannon Yang, Thomas Henighan,
  Timothy Maxwell, Timothy Telleen-Lawton, Tristan Hume, Zac Hatfield-Dodds,
  Jared Kaplan, Jan Brauner, Samuel~R. Bowman, and Ethan Perez.
\newblock Measuring faithfulness in chain-of-thought reasoning, 2023.
\newblock URL \url{https://arxiv.org/abs/2307.13702}.

\bibitem[Li et~al.(2026{\natexlab{a}})Li, Zhang, Zhao, Wang, and
  Sun]{li2026defertoplan}
Nuoran Li, Zhang Zhang, Yueran Zhao, Tianze Wang, and Chao Sun.
\newblock Defer to plan: Adaptive multi-agent fusion for end-to-end {V2X}
  driving, 2026{\natexlab{a}}.
\newblock URL \url{https://arxiv.org/abs/2607.19774}.
\newblock Accepted at IEEE ICME 2026.

\bibitem[Li et~al.(2023)Li, Du, Zhou, Wang, Zhao, and Wen]{li2023pope}
Yifan Li, Yifan Du, Kun Zhou, Jinpeng Wang, Wayne~Xin Zhao, and Ji-Rong Wen.
\newblock Evaluating object hallucination in large vision-language models.
\newblock In \emph{Proceedings of the 2023 Conference on Empirical Methods in
  Natural Language Processing}, pp.\  292--305. Association for Computational
  Linguistics, 2023.
\newblock \doi{10.18653/v1/2023.emnlp-main.20}.
\newblock URL \url{https://aclanthology.org/2023.emnlp-main.20/}.

\bibitem[Li et~al.(2025)Li, Xiong, Guo, Li, Yan, Xu, Zhou, Chen, Sun, Wang, Ma,
  Chen, Ye, Liu, and Wang]{li2025recogdrive}
Yongkang Li, Kaixin Xiong, Xiangyu Guo, Fang Li, Sixu Yan, Gangwei Xu, Lijun
  Zhou, Long Chen, Haiyang Sun, Bing Wang, Kun Ma, Guang Chen, Hangjun Ye,
  Wenyu Liu, and Xinggang Wang.
\newblock {ReCogDrive}: A reinforced cognitive framework for end-to-end
  autonomous driving.
\newblock \emph{arXiv preprint arXiv:2506.08052}, 2025.
\newblock URL \url{https://arxiv.org/abs/2506.08052}.

\bibitem[Li et~al.(2026{\natexlab{b}})Li, Tian, Zhu, Zhu, Lin, Xiong, and
  Zhao]{li2025driver1}
Yue Li, Meng Tian, Dechang Zhu, Jiangtong Zhu, Zhenyu Lin, Zhiwei Xiong, and
  Xinhai Zhao.
\newblock {Drive-R1}: Bridging reasoning and planning in {VLMs} for autonomous
  driving with reinforcement learning.
\newblock In \emph{Proceedings of the AAAI Conference on Artificial
  Intelligence}, volume~40, pp.\  6708--6716, 2026{\natexlab{b}}.
\newblock \doi{10.1609/aaai.v40i8.37602}.
\newblock URL \url{https://ojs.aaai.org/index.php/AAAI/article/view/37602}.

\bibitem[Liao et~al.(2025)Liao, Qi, Shu, Zhang, Lin, Liu, and
  Ma]{liao2025robodrivevlm}
Dacheng Liao, Mengshi Qi, Peng Shu, Zhining Zhang, Yuxin Lin, Liang Liu, and
  Huadong Ma.
\newblock {RoboDriveVLM}: A novel benchmark and baseline towards robust
  vision-language models for autonomous driving.
\newblock \emph{arXiv preprint arXiv:2512.01300}, 2025.
\newblock URL \url{https://arxiv.org/abs/2512.01300}.

\bibitem[Liu et~al.(2025{\natexlab{a}})Liu, Liu, Wang, Yang, and
  Chen]{liu2025colmdriver}
Changxing Liu, Genjia Liu, Zijun Wang, Jinchang Yang, and Siheng Chen.
\newblock {CoLMDriver}: {LLM}-based negotiation benefits cooperative autonomous
  driving, 2025{\natexlab{a}}.
\newblock URL \url{https://arxiv.org/abs/2503.08683}.

\bibitem[Liu et~al.(2026)Liu, Li, Wang, Zhang, and He]{liu2026anchorvla}
Qi~Liu, Yabei Li, Hongsong Wang, Heng Zhang, and Lei He.
\newblock {AnchorVLA}: Bridging discrete decisions and continuous trajectories
  for vision-language-action planning, 2026.
\newblock URL \url{https://arxiv.org/abs/2607.03182}.

\bibitem[Liu et~al.(2025{\natexlab{b}})Liu, Huang, Yang, Yan, Wang, Hou, Lin,
  Bai, and Zhao]{liu2026drivepi}
Zhe Liu, Runhui Huang, Rui Yang, Siming Yan, Zining Wang, Lu~Hou, Di~Lin, Xiang
  Bai, and Hengshuang Zhao.
\newblock {DrivePI}: Spatial-aware {4D} {MLLM} for unified autonomous driving
  understanding, perception, prediction and planning, 2025{\natexlab{b}}.
\newblock URL \url{https://arxiv.org/abs/2512.12799}.

\bibitem[Lu et~al.(2026)Lu, Guan, Huang, Li, Li, Kong, Li, Wang, Xu, Luo, Li,
  Dang, Wang, Xu, Wu, Wu, Hao, Zhang, Jiang, Zhang, Zhou, Tang, Wang, Gao, Bu,
  Tian, Qiu, Jia, Liu, Ge, Li, Shen, Cui, Xie, Wang, Sun, Zhao, Huang, Liu,
  Zhu, Jiang, Guo, Gong, Leng, Ma, Wang, Chen, Yang, Ye, and Chen]{lu2026onevl}
Jinghui Lu, Jiayi Guan, Zhijian Huang, Jinlong Li, Guang Li, Lingdong Kong,
  Yingyan Li, Han Wang, Shaoqing Xu, Yuechen Luo, Fang Li, Chenxu Dang, Junli
  Wang, Tao Xu, Jing Wu, Jianhua Wu, Xiaoshuai Hao, Wen Zhang, Tianyi Jiang,
  Lingfeng Zhang, Lei Zhou, Yingbo Tang, Jie Wang, Yinfeng Gao, Xizhou Bu,
  Haochen Tian, Yihang Qiu, Feiyang Jia, Lin Liu, Yigu Ge, Hanbing Li, Yuannan
  Shen, Jianwei Cui, Hongwei Xie, Bing Wang, Haiyang Sun, Jingwei Zhao, Jiahui
  Huang, Pei Liu, Zeyu Zhu, Yuncheng Jiang, Zibin Guo, Chuhong Gong, Hanchao
  Leng, Kun Ma, Naiyan Wang, Guang Chen, Kuiyuan Yang, Hangjun Ye, and Long
  Chen.
\newblock {Xiaomi OneVL}: One-step latent reasoning and planning with
  vision-language explanation.
\newblock \emph{arXiv preprint arXiv:2604.18486}, 2026.
\newblock URL \url{https://arxiv.org/abs/2604.18486}.

\bibitem[Lu et~al.(2024)Lu, Hu, Zhong, Wang, Wang, and Chen]{HEAL}
Yifan Lu, Yue Hu, Yiqi Zhong, Dequan Wang, Yanfeng Wang, and Siheng Chen.
\newblock An extensible framework for open heterogeneous collaborative
  perception.
\newblock \emph{arXiv preprint arXiv:2401.13964}, 2024.
\newblock URL \url{https://arxiv.org/abs/2401.13964}.

\bibitem[Luo et~al.(2026)Luo, Yang, Fan, Gao, Yu, Li, Tu, Zhou, and
  Liu]{luo2026v2xunipool}
Xuewen Luo, Fengze Yang, Ding Fan, Xiangbo Gao, Bo~Yu, Zihao Li, Zhengzhong Tu,
  Yang Zhou, and Chenxi Liu.
\newblock {V2X-UniPool}: Unifying multimodal perception and knowledge reasoning
  for autonomous driving.
\newblock In \emph{Proceedings of the IEEE/CVF Conference on Computer Vision
  and Pattern Recognition (CVPR) Workshops}, pp.\  747--756, June 2026.
\newblock URL
  \url{https://openaccess.thecvf.com/content/CVPR2026W/DriveX/html/Luo_V2X-UniPool_Unifying_Multimodal_Perception_and_Knowledge_Reasoning_for_Autonomous_Driving_CVPRW_2026_paper.html}.

\bibitem[Marcu et~al.(2023)Marcu, Chen, H{\"u}nermann, Karnsund, Hanotte,
  Chidananda, Nair, Badrinarayanan, Kendall, Shotton, Arani, and
  Sinavski]{lingoqa}
Ana-Maria Marcu, Long Chen, Jan H{\"u}nermann, Alice Karnsund, Benoit Hanotte,
  Prajwal Chidananda, Saurabh Nair, Vijay Badrinarayanan, Alex Kendall, Jamie
  Shotton, Elahe Arani, and Oleg Sinavski.
\newblock {LingoQA}: Visual question answering for autonomous driving.
\newblock \emph{arXiv preprint arXiv:2312.14115}, 2023.

\bibitem[Nie et~al.(2024)Nie, Peng, Wang, Cai, Han, Xu, and
  Zhang]{reason2drive}
Ming Nie, Renyuan Peng, Chunwei Wang, Xinyue Cai, Jianhua Han, Hang Xu, and
  Li~Zhang.
\newblock {Reason2Drive}: Towards interpretable and chain-based reasoning for
  autonomous driving.
\newblock In \emph{European Conference on Computer Vision}, pp.\  292--308,
  2024.
\newblock \doi{10.1007/978-3-031-73347-5_17}.

\bibitem[Pan et~al.(2024)Pan, Yaman, Nesti, Mallik, Allievi, Velipasalar, and
  Ren]{vlp}
Chenbin Pan, Burhaneddin Yaman, Tommaso Nesti, Abhirup Mallik, Alessandro~G.
  Allievi, Senem Velipasalar, and Liu Ren.
\newblock {VLP}: Vision language planning for autonomous driving.
\newblock \emph{arXiv preprint arXiv:2401.05577}, 2024.

\bibitem[Panda et~al.(2026)Panda, Maia, Agarwal, and Greer]{panda2026cvaa}
Kalpana Panda, Wesley Maia, Vinti Agarwal, and Ross Greer.
\newblock What do they see? interpreting complex road scenarios through the
  eyes of vision-language-action models for safe and trustworthy autonomous
  vehicle learning, 2026.
\newblock URL \url{https://arxiv.org/abs/2607.16938}.

\bibitem[Peng et~al.(2026)Peng, Lu, Zhou, Cui, Chen, and Wang]{peng2026omniv2x}
Juntong Peng, Juanwu Lu, Yupeng Zhou, Can Cui, Yaobin Chen, and Ziran Wang.
\newblock {OmniV2X}: A generative foundation planner for efficient end-to-end
  cooperative driving, 2026.
\newblock URL \url{https://arxiv.org/abs/2606.21165}.

\bibitem[Qian et~al.(2023)Qian, Chen, Zhuo, Jiao, and Jiang]{nuscenesqa}
Tianwen Qian, Jingjing Chen, Linhai Zhuo, Yang Jiao, and Yu-Gang Jiang.
\newblock {NuScenes-QA}: A multi-modal visual question answering benchmark for
  autonomous driving scenario.
\newblock \emph{arXiv preprint arXiv:2305.14836}, 2023.

\bibitem[Renz et~al.(2025)Renz, Chen, Arani, and Sinavski]{simlingo}
Katrin Renz, Long Chen, Elahe Arani, and Oleg Sinavski.
\newblock {SimLingo}: Vision-only closed-loop autonomous driving with
  language-action alignment.
\newblock In \emph{Proceedings of the IEEE/CVF Conference on Computer Vision
  and Pattern Recognition}, pp.\  11993--12003, 2025.
\newblock \doi{10.1109/CVPR52734.2025.01120}.
\newblock URL
  \url{https://openaccess.thecvf.com/content/CVPR2025/html/Renz_SimLingo_Vision-Only_Closed-Loop_Autonomous_Driving_with_Language-Action_Alignment_CVPR_2025_paper.html}.

\bibitem[Richard et~al.(2026)Richard, Varghese, Pham, Oh, and
  Das]{richard2026d2v2x}
Kevin Richard, Alphin Varghese, Colin Pham, David Oh, and Srijan Das.
\newblock {D2-V2X}: Depth-driven cooperative {V2X} reasoning for autonomous
  driving, 2026.
\newblock URL \url{https://arxiv.org/abs/2605.24098}.

\bibitem[Rochman et~al.(2023)Rochman, Ye, Zhang, and Ghosh]{rochman2023eval}
Muhammad~Iqbal Rochman, Wei Ye, Zhi-Li Zhang, and Monisha Ghosh.
\newblock A comprehensive real-world evaluation of {5G} improvements over {4G}
  in low- and mid-bands.
\newblock \emph{arXiv preprint arXiv:2312.00957}, 2023.
\newblock URL \url{https://arxiv.org/abs/2312.00957}.

\bibitem[Rohrbach et~al.(2018)Rohrbach, Hendricks, Burns, Darrell, and
  Saenko]{rohrbach2018chair}
Anna Rohrbach, Lisa~Anne Hendricks, Kaylee Burns, Trevor Darrell, and Kate
  Saenko.
\newblock Object hallucination in image captioning.
\newblock In \emph{Proceedings of the 2018 Conference on Empirical Methods in
  Natural Language Processing}, pp.\  4035--4045. Association for Computational
  Linguistics, 2018.
\newblock \doi{10.18653/v1/D18-1437}.
\newblock URL \url{https://aclanthology.org/D18-1437/}.

\bibitem[Shah et~al.(2026)Shah, Sharan, Goel, Pasula, Hebbalae, Choi, and
  Chinchali]{shah2026crossview}
Sahil Shah, S~P Sharan, Harsh Goel, Manvik Pasula, Adithya Hebbalae, Minkyu
  Choi, and Sandeep~P. Chinchali.
\newblock {CrossView}: Can vision-language models reason across cameras?, 2026.
\newblock URL \url{https://arxiv.org/abs/2608.15539}.
\newblock ECCV 2026.

\bibitem[Shao et~al.(2023)Shao, Hu, Wang, Waslander, Liu, and Li]{lmdrive}
Hao Shao, Yuxuan Hu, Letian Wang, Steven~L. Waslander, Yu~Liu, and Hongsheng
  Li.
\newblock {LMDrive}: Closed-loop end-to-end driving with large language models.
\newblock \emph{arXiv preprint arXiv:2312.07488}, 2023.

\bibitem[Sima et~al.(2024)Sima, Renz, Chitta, Chen, Zhang, Xie, Bei{\ss}wenger,
  Luo, Geiger, and Li]{drivelm}
Chonghao Sima, Katrin Renz, Kashyap Chitta, Li~Chen, Hanxue Zhang, Chengen Xie,
  Jens Bei{\ss}wenger, Ping Luo, Andreas Geiger, and Hongyang Li.
\newblock {DriveLM}: Driving with graph visual question answering.
\newblock In \emph{European Conference on Computer Vision}, pp.\  256--274.
  Springer, 2024.
\newblock \doi{10.1007/978-3-031-72943-0_15}.

\bibitem[Song et~al.(2025)Song, Yang, Wen, and Li]{Traf-align}
Zhiying Song, Lei Yang, Fuxi Wen, and Jun Li.
\newblock {TraF-Align}: Trajectory-aware feature alignment for asynchronous
  multi-agent perception.
\newblock In \emph{Proceedings of the Computer Vision and Pattern Recognition
  Conference}, pp.\  12048--12057, 2025.

\bibitem[Song et~al.(2026{\natexlab{a}})Song, Xia, Wang, Yu, Zhou, and
  Niu]{song2025unimmv2x}
Ziyi Song, Chen Xia, Chenbing Wang, Haibao Yu, Sheng Zhou, and Zhisheng Niu.
\newblock {UniMM-V2X}: {MoE}-enhanced multi-level fusion for end-to-end
  cooperative autonomous driving.
\newblock In \emph{Proceedings of the AAAI Conference on Artificial
  Intelligence}, volume~40, pp.\  9135--9143, 2026{\natexlab{a}}.
\newblock \doi{10.1609/aaai.v40i11.37870}.
\newblock URL \url{https://ojs.aaai.org/index.php/AAAI/article/view/37870}.

\bibitem[Song et~al.(2026{\natexlab{b}})Song, Xia, Yu, Zhou, and
  Niu]{song2026dhvlm}
Ziyi Song, Chen Xia, Hang Yu, Sheng Zhou, and Zhisheng Niu.
\newblock {DH-VLM}: Dual-horizon cooperative latent reasoning for autonomous
  driving, 2026{\natexlab{b}}.
\newblock URL \url{https://arxiv.org/abs/2608.09333}.

\bibitem[Tian et~al.(2026)Tian, Lian, Yang, Chen, and Li]{tian2026ccot}
Kefei Tian, Yuansheng Lian, Kai Yang, Xiangdong Chen, and Shen Li.
\newblock {C-CoT}: Counterfactual chain-of-thought with vision-language models
  for safe autonomous driving, 2026.
\newblock URL \url{https://arxiv.org/abs/2605.10744}.

\bibitem[Turpin et~al.(2023)Turpin, Michael, Perez, and
  Bowman]{turpin2023unfaithful}
Miles Turpin, Julian Michael, Ethan Perez, and Samuel~R. Bowman.
\newblock Language models don't always say what they think: Unfaithful
  explanations in chain-of-thought prompting.
\newblock In \emph{Advances in Neural Information Processing Systems},
  volume~36, pp.\  74952--74965, 2023.
\newblock URL \url{https://arxiv.org/abs/2305.04388}.

\bibitem[Vo et~al.(2026)Vo, Vo, Nguyen, Tran, Nguyen, Cuong, Gawugah,
  Godavarthi, Rainwater, Bui, Nguyen, Nguyen, and Le]{vo2026drivespatial}
Hao Vo, Khoa Vo, Phu~Loc Nguyen, Sieu Tran, Duc~Minh Nguyen, Ngo~Xuan Cuong,
  Gladys Gawugah, Sreevenkata Anjani~Tishita Godavarthi, Chase Rainwater, Nghi
  D.~Q. Bui, Anh Nguyen, Duy Minh~Ho Nguyen, and Ngan Le.
\newblock {DriveSpatial}: A benchmark for spatiotemporal intelligence in {VLMs}
  for autonomous driving, 2026.
\newblock URL \url{https://arxiv.org/abs/2605.23176}.

\bibitem[Wang et~al.(2026{\natexlab{a}})Wang, Tang, Ren, Zhao, Feng, and
  Ma]{wang2026vlaworld}
Guoqing Wang, Pin Tang, Xiangxuan Ren, Guodongfang Zhao, Bailan Feng, and Chao
  Ma.
\newblock Learning vision-language-action world models for autonomous driving,
  2026{\natexlab{a}}.
\newblock URL \url{https://arxiv.org/abs/2604.09059}.

\bibitem[Wang et~al.(2026{\natexlab{b}})Wang, Li, Huang, Dang, Ye, Han, and
  Chen]{wang2026vggdrive}
Jie Wang, Guang Li, Zhijian Huang, Chenxu Dang, Hangjun Ye, Yahong Han, and
  Long Chen.
\newblock {VGGDrive}: Empowering vision-language models with cross-view
  geometric grounding for autonomous driving, 2026{\natexlab{b}}.
\newblock URL \url{https://arxiv.org/abs/2602.20794}.

\bibitem[Wang et~al.(2025)Wang, Yu, Jiang, Lan, Shi, Chang, Kautz, Li, and
  Alvarez]{wang2025omnidrive}
Shihao Wang, Zhiding Yu, Xiaohui Jiang, Shiyi Lan, Min Shi, Nadine Chang, Jan
  Kautz, Ying Li, and Jose~M. Alvarez.
\newblock {OmniDrive}: A holistic vision-language dataset for autonomous
  driving with counterfactual reasoning.
\newblock In \emph{Proceedings of the IEEE/CVF Conference on Computer Vision
  and Pattern Recognition}, pp.\  22442--22452, June 2025.
\newblock \doi{10.1109/CVPR52734.2025.02090}.
\newblock URL
  \url{https://openaccess.thecvf.com/content/CVPR2025/html/Wang_OmniDrive_A_Holistic_Vision-Language_Dataset_for_Autonomous_Driving_with_Counterfactual_CVPR_2025_paper.html}.

\bibitem[Wang et~al.(2020)Wang, Manivasagam, Liang, Yang, Zeng, and
  Urtasun]{v2vnet}
Tsun-Hsuan Wang, Sivabalan Manivasagam, Ming Liang, Bin Yang, Wenyuan Zeng, and
  Raquel Urtasun.
\newblock {V2VNet}: Vehicle-to-vehicle communication for joint perception and
  prediction.
\newblock In \emph{Computer Vision--ECCV 2020: 16th European Conference,
  Glasgow, UK, August 23--28, 2020, Proceedings, Part II 16}, pp.\  605--621.
  Springer, 2020.
\newblock \doi{10.1007/978-3-030-58536-5_36}.

\bibitem[Wei et~al.(2023)Wei, Wei, Hu, Lu, Zhong, Chen, and Zhang]{CoBEVFlow}
Sizhe Wei, Yuxi Wei, Yue Hu, Yifan Lu, Yiqi Zhong, Siheng Chen, and Ya~Zhang.
\newblock Asynchrony-robust collaborative perception via bird's eye view flow.
\newblock In \emph{Advances in Neural Information Processing Systems},
  volume~36, pp.\  28462--28477, 2023.

\bibitem[Xia et~al.(2025)Xia, Yuan, Luo, Fu, Li, Zhu, Luo, Chen, and
  Li]{PolyInter}
Yuchen Xia, Quan Yuan, Guiyang Luo, Xiaoyuan Fu, Yang Li, Xuanhan Zhu, Tianyou
  Luo, Siheng Chen, and Jinglin Li.
\newblock One is plenty: A polymorphic feature interpreter for immutable
  heterogeneous collaborative perception.
\newblock In \emph{Proceedings of the Computer Vision and Pattern Recognition
  Conference}, pp.\  1592--1601, 2025.
\newblock \doi{10.1109/CVPR52734.2025.00156}.

\bibitem[Xiang et~al.(2024)Xiang, Zheng, Xia, Xu, Gao, Zhou, Han, Ji, Li, Meng,
  Jin, Lei, Ma, He, Ma, Yuan, Zhao, and Ma]{v2xreal}
Hao Xiang, Zhaoliang Zheng, Xin Xia, Runsheng Xu, Letian Gao, Zewei Zhou,
  Xu~Han, Xinkai Ji, Mingxi Li, Zonglin Meng, Li~Jin, Mingyue Lei, Zhaoyang Ma,
  Zihang He, Haoxuan Ma, Yunshuang Yuan, Yingqian Zhao, and Jiaqi Ma.
\newblock {V2X-Real}: A large-scale dataset for vehicle-to-everything
  cooperative perception.
\newblock In \emph{European Conference on Computer Vision}, pp.\  455--470.
  Springer, 2024.

\bibitem[Xie et~al.(2025)Xie, Kong, Dong, Sima, Zhang, Chen, Liu, and
  Pan]{xie2025drivebench}
Shaoyuan Xie, Lingdong Kong, Yuhao Dong, Chonghao Sima, Wenwei Zhang, Qi~Alfred
  Chen, Ziwei Liu, and Liang Pan.
\newblock Are {VLMs} ready for autonomous driving? an empirical study from the
  reliability, data and metric perspectives.
\newblock In \emph{Proceedings of the IEEE/CVF International Conference on
  Computer Vision}, pp.\  6585--6597, October 2025.
\newblock \doi{10.1109/ICCV51701.2025.00621}.
\newblock URL
  \url{https://openaccess.thecvf.com/content/ICCV2025/html/Xie_Are_VLMs_Ready_for_Autonomous_Driving_An_Empirical_Study_from_ICCV_2025_paper.html}.

\bibitem[Xu et~al.(2022{\natexlab{a}})Xu, Xiang, Tu, Xia, Yang, and
  Ma]{v2x-vit}
Runsheng Xu, Hao Xiang, Zhengzhong Tu, Xin Xia, Ming-Hsuan Yang, and Jiaqi Ma.
\newblock {V2X-ViT}: Vehicle-to-everything cooperative perception with vision
  transformer.
\newblock In \emph{Proceedings of the European Conference on Computer Vision},
  pp.\  107--124, 2022{\natexlab{a}}.
\newblock \doi{10.1007/978-3-031-19842-7_7}.

\bibitem[Xu et~al.(2022{\natexlab{b}})Xu, Xiang, Xia, Han, Li, and Ma]{OPV2V}
Runsheng Xu, Hao Xiang, Xin Xia, Xu~Han, Jinlong Li, and Jiaqi Ma.
\newblock {OPV2V}: An open benchmark dataset and fusion pipeline for perception
  with vehicle-to-vehicle communication.
\newblock In \emph{2022 International Conference on Robotics and Automation
  (ICRA)}, pp.\  2583--2589. IEEE, 2022{\natexlab{b}}.
\newblock \doi{10.1109/ICRA46639.2022.9812038}.

\bibitem[Xu et~al.(2023{\natexlab{a}})Xu, Tu, Xiang, Shao, Zhou, and
  Ma]{CoBEVT}
Runsheng Xu, Zhengzhong Tu, Hao Xiang, Wei Shao, Bolei Zhou, and Jiaqi Ma.
\newblock {CoBEVT}: Cooperative bird's eye view semantic segmentation with
  sparse transformers.
\newblock In \emph{Proceedings of the 6th Conference on Robot Learning}, volume
  205 of \emph{Proceedings of Machine Learning Research}, pp.\  989--1000.
  PMLR, 2023{\natexlab{a}}.
\newblock URL \url{https://proceedings.mlr.press/v205/xu23a.html}.

\bibitem[Xu et~al.(2026)Xu, Wu, Wu, Xu, Jiang, Wang, Ke, Jiang, Zhang, and
  Wang]{xu2026roadsidecooperative}
Yitao Xu, Tong Wu, Yiyan Wu, Guoji Xu, Yanbo Jiang, Jiahao Wang, Zehong Ke,
  Junkai Jiang, Fang Zhang, and Jianqiang Wang.
\newblock Roadside-cooperative autonomous driving: From data platform to
  vision-language end-to-end reasoning, 2026.
\newblock URL \url{https://arxiv.org/abs/2608.21032}.

\bibitem[Xu et~al.(2023{\natexlab{b}})Xu, Zhang, Xie, Zhao, Guo, Wong, Li, and
  Zhao]{drivegpt4}
Zhenhua Xu, Yujia Zhang, Enze Xie, Zhen Zhao, Yong Guo, Kwan-Yee~K. Wong,
  Zhenguo Li, and Hengshuang Zhao.
\newblock {DriveGPT4}: Interpretable end-to-end autonomous driving via large
  language model.
\newblock \emph{arXiv preprint arXiv:2310.01412}, 2023{\natexlab{b}}.

\bibitem[Yang et~al.(2023)Yang, Yang, Wang, Liu, Xu, Yin, Zhai, and
  Zhang]{how2comm}
Dingkang Yang, Kun Yang, Yuzheng Wang, Jing Liu, Zhi Xu, Rongbin Yin, Peng
  Zhai, and Lihua Zhang.
\newblock {How2comm}: Communication-efficient and collaboration-pragmatic
  multi-agent perception.
\newblock In \emph{Advances in Neural Information Processing Systems},
  volume~36, pp.\  25151--25164, 2023.
\newblock \doi{10.52202/075280-1093}.

\bibitem[Yang et~al.(2025{\natexlab{a}})Yang, Bu, Li, Li, Wang, and Li]{ACCO}
Kang Yang, Tianci Bu, Lantao Li, Chunxu Li, Yongcai Wang, and Deying Li.
\newblock Is discretization fusion all you need for collaborative perception?
\newblock In \emph{2025 IEEE International Conference on Robotics and
  Automation (ICRA)}, pp.\  9590--9596, 2025{\natexlab{a}}.
\newblock \doi{10.1109/ICRA55743.2025.11128776}.

\bibitem[Yang et~al.(2026{\natexlab{a}})Yang, Bu, Wang, Li, Jie, and
  Wang]{yang2026uecp}
Kang Yang, Tianci Bu, Peng Wang, Deying Li, Wen Jie, and Yongcai Wang.
\newblock {UECP}: Uncertainty-enhanced collaborative perception,
  2026{\natexlab{a}}.
\newblock URL \url{https://arxiv.org/abs/2606.23046}.

\bibitem[Yang et~al.(2026{\natexlab{b}})Yang, Bu, Wang, Li, and
  Wang]{yang2026bolt}
Kang Yang, Tianci Bu, Peng Wang, Deying Li, and Yongcai Wang.
\newblock {BOLT}: Online lightweight adaptation for preparation-free
  heterogeneous cooperative perception, 2026{\natexlab{b}}.
\newblock URL \url{https://arxiv.org/abs/2605.00405}.

\bibitem[Yang et~al.(2026{\natexlab{c}})Yang, Wang, Li, Bu, Sun, Li, and
  Wang]{EIMC}
Kang Yang, Peng Wang, Lantao Li, Tianci Bu, Chen Sun, Deying Li, and Yongcai
  Wang.
\newblock {EIMC}: Efficient instance-aware multi-modal collaborative
  perception.
\newblock \emph{arXiv preprint arXiv:2603.02532}, 2026{\natexlab{c}}.
\newblock URL \url{https://arxiv.org/abs/2603.02532}.

\bibitem[Yang et~al.(2025{\natexlab{b}})Yang, Huang, Yin, Hu, Xu, Luo, Sun, Wu,
  Ao, Zhu, Wen, and Wang]{yang2025monirefer}
Panquan Yang, Junfei Huang, Zongzhangbao Yin, Yingsong Hu, Anni Xu, Xinyi Luo,
  Xueqi Sun, Hai Wu, Sheng Ao, Zhaoxing Zhu, Chenglu Wen, and Cheng Wang.
\newblock {MoniRefer}: A real-world large-scale multi-modal dataset based on
  roadside infrastructure for {3D} visual grounding, 2025{\natexlab{b}}.
\newblock URL \url{https://arxiv.org/abs/2512.24605}.

\bibitem[Yang et~al.(2026{\natexlab{d}})Yang, Chai, Jia, Li, Shao, Zhu, Su, and
  Yan]{yang2026drivemoe}
Zhenjie Yang, Yilin Chai, Xiaosong Jia, Qifeng Li, Yuqian Shao, Xuekai Zhu,
  Haisheng Su, and Junchi Yan.
\newblock {DriveMoE}: Mixture-of-experts for vision-language-action model in
  end-to-end autonomous driving.
\newblock In \emph{Proceedings of the IEEE/CVF Conference on Computer Vision
  and Pattern Recognition (CVPR)}, pp.\  10678--10688, June 2026{\natexlab{d}}.
\newblock URL
  \url{https://openaccess.thecvf.com/content/CVPR2026/html/Yang_DriveMoE_Mixture-of-Experts_for_Vision-Language-Action_Model_in_End-to-End_Autonomous_Driving_CVPR_2026_paper.html}.

\bibitem[Yin et~al.(2025)Yin, Kan, and Watzenig]{yin2025map}
Huilin Yin, Yiming Kan, and Daniel Watzenig.
\newblock {MAP}: End-to-end autonomous driving with map-assisted planning,
  2025.
\newblock URL \url{https://arxiv.org/abs/2509.13926}.

\bibitem[You et~al.(2024)You, Shi, Jiang, Huang, Gan, Wu, Cheng, Li, and
  Ran]{you2024v2xvlm}
Junwei You, Haotian Shi, Zhuoyu Jiang, Zilin Huang, Rui Gan, Keshu Wu,
  Xi~Cheng, Xiaopeng Li, and Bin Ran.
\newblock {V2X-VLM}: End-to-end {V2X} cooperative autonomous driving through
  large vision-language models, 2024.
\newblock URL \url{https://arxiv.org/abs/2408.09251}.

\bibitem[You et~al.(2025)You, Li, Jiang, Huang, Gan, Shi, and Ran]{you2025seal}
Junwei You, Pei Li, Zhuoyu Jiang, Zilin Huang, Rui Gan, Haotian Shi, and Bin
  Ran.
\newblock {SEAL}: Vision-language model-based safe end-to-end cooperative
  autonomous driving with adaptive long-tail modeling, 2025.
\newblock URL \url{https://arxiv.org/abs/2506.21041}.

\bibitem[You et~al.(2026)You, Li, Jiang, Tang, Huang, Gan, Liu, Zhao, Chen, and
  Ran]{you2026v2xqa}
Junwei You, Pei Li, Zhuoyu Jiang, Weizhe Tang, Zilin Huang, Rui Gan, Jiaxi Liu,
  Yan Zhao, Sikai Chen, and Bin Ran.
\newblock {V2X-QA}: A comprehensive reasoning dataset and benchmark for
  multimodal large language models in autonomous driving across ego,
  infrastructure, and cooperative views, 2026.
\newblock URL \url{https://arxiv.org/abs/2604.02710}.

\bibitem[Yu et~al.(2022)Yu, Luo, Shu, Huo, Yang, Shi, Guo, Li, Hu, Yuan, and
  Nie]{dair-v2x}
Haibao Yu, Yizhen Luo, Mao Shu, Yiyi Huo, Zebang Yang, Yifeng Shi, Zhenglong
  Guo, Hanyu Li, Xing Hu, Jirui Yuan, and Zaiqing Nie.
\newblock {DAIR-V2X}: A large-scale dataset for vehicle-infrastructure
  cooperative {3D} object detection.
\newblock In \emph{Proceedings of the IEEE/CVF Conference on Computer Vision
  and Pattern Recognition}, pp.\  21361--21370, 2022.
\newblock \doi{10.1109/CVPR52688.2022.02067}.

\bibitem[Yu et~al.(2023)Yu, Yang, Ruan, Yang, Tang, Gao, Hao, Shi, Pan, Sun,
  Song, Yuan, Luo, and Nie]{v2x-seq}
Haibao Yu, Wenxian Yang, Hongzhi Ruan, Zhenwei Yang, Yingjuan Tang, Xu~Gao, Xin
  Hao, Yifeng Shi, Yifeng Pan, Ning Sun, Juan Song, Jirui Yuan, Ping Luo, and
  Zaiqing Nie.
\newblock {V2X-Seq}: A large-scale sequential dataset for
  vehicle-infrastructure cooperative perception and forecasting.
\newblock In \emph{Proceedings of the IEEE/CVF Conference on Computer Vision
  and Pattern Recognition}, pp.\  5486--5495, 2023.
\newblock \doi{10.1109/CVPR52729.2023.00531}.

\bibitem[Yu et~al.(2025)Yu, Yang, Zhong, Yang, Fan, Luo, and Nie]{yu2025univ2x}
Haibao Yu, Wenxian Yang, Jiaru Zhong, Zhenwei Yang, Siqi Fan, Ping Luo, and
  Zaiqing Nie.
\newblock End-to-end autonomous driving through {V2X} cooperation.
\newblock In \emph{Proceedings of the AAAI Conference on Artificial
  Intelligence}, volume~39, pp.\  9598--9606, 2025.
\newblock \doi{10.1609/aaai.v39i9.33040}.
\newblock URL \url{https://ojs.aaai.org/index.php/AAAI/article/view/33040}.

\bibitem[Yu et~al.(2026)Yu, Wu, Xu, and Ma]{yu2026hyworldvla}
Quanfu Yu, Xian Wu, Hao Xu, and Liulong Ma.
\newblock {HyWorldVLA}: A vision-language-action model with hybrid world
  modeling for autonomous driving, 2026.
\newblock URL \url{https://arxiv.org/abs/2607.20988}.

\bibitem[Yuan et~al.(2025)Yuan, Xia, Cremers, and Sester]{sparsealign}
Yunshuang Yuan, Yan Xia, Daniel Cremers, and Monika Sester.
\newblock {SparseAlign}: a fully sparse framework for cooperative object
  detection.
\newblock In \emph{Proceedings of the Computer Vision and Pattern Recognition
  Conference}, pp.\  22296--22305, 2025.
\newblock \doi{10.1109/CVPR52734.2025.02077}.

\bibitem[Zhang et~al.(2025)Zhang, Yuan, Chen, Liao, Chen, Shen, Zhou, and
  Chua]{zhang2025reasoningvla}
Dapeng Zhang, Zhenlong Yuan, Zhangquan Chen, Chih-Ting Liao, Yinda Chen, Fei
  Shen, Qingguo Zhou, and Tat-Seng Chua.
\newblock {Reasoning-VLA}: A fast and general vision-language-action reasoning
  model for autonomous driving.
\newblock \emph{arXiv preprint arXiv:2511.19912}, 2025.
\newblock URL \url{https://arxiv.org/abs/2511.19912}.

\bibitem[Zhang et~al.(2026)Zhang, Chen, Gao, Wang, Ge, Hu, Shi, and
  Zhang]{zhang2026onedrive}
Yiwei Zhang, Xuesong Chen, Jin Gao, Hanshi Wang, Fudong Ge, Weiming Hu,
  Shaoshuai Shi, and Zhipeng Zhang.
\newblock {OneDrive}: Unified multi-paradigm driving with
  vision-language-action models, 2026.
\newblock URL \url{https://arxiv.org/abs/2604.17915}.

\bibitem[Zheng et~al.(2026)Zheng, Huang, Sun, Li, and Zhao]{zheng2026drivema}
Weicheng Zheng, Yixin Huang, Qiao Sun, Derun Li, and Hang Zhao.
\newblock {DriveMA}: Driving vision-language-action models with verifiable
  meta-actions, 2026.
\newblock URL \url{https://arxiv.org/abs/2605.31271}.

\bibitem[Zhou et~al.(2025{\natexlab{a}})Zhou, Larintzakis, Guo, Zimmer, Liu,
  Cao, Zhang, Lakshminarasimhan, Strand, and Knoll]{zhou2025tumtrafficvideoqa}
Xingcheng Zhou, Konstantinos Larintzakis, Hao Guo, Walter Zimmer, Mingyu Liu,
  Hu~Cao, Jiajie Zhang, Venkatnarayanan Lakshminarasimhan, Leah Strand, and
  Alois~C. Knoll.
\newblock {TUMTraffic-VideoQA}: A benchmark for unified spatio-temporal video
  understanding in traffic scenes, 2025{\natexlab{a}}.
\newblock URL \url{https://arxiv.org/abs/2502.02449}.

\bibitem[Zhou et~al.(2025{\natexlab{b}})Zhou, Cai, Zhao, Zhang, Huang, Zhou,
  and Ma]{zhou2025autovla}
Zewei Zhou, Tianhui Cai, Seth~Z. Zhao, Yun Zhang, Zhiyu Huang, Bolei Zhou, and
  Jiaqi Ma.
\newblock {AutoVLA}: A vision-language-action model for end-to-end autonomous
  driving with adaptive reasoning and reinforcement fine-tuning,
  2025{\natexlab{b}}.
\newblock URL \url{https://arxiv.org/abs/2506.13757}.

\end{thebibliography}
\bibliographystyle{iclr2027_conference}

\clearpage
\appendix
\raggedbottom
\renewcommand{\floatpagefraction}{0.8}

\noindent\textbf{Appendix guide.}
Appendices~\ref{sec:appendix-annotation-validation} and~\ref{sec:appendix-prompts}
document benchmark construction, validation, and templates.
Appendix~\ref{sec:appendix-reproducibility-release} specifies training and evaluation.
Supplementary CDQA results, supervision ablations, and roadside-input analyses
appear in Appendices~\ref{sec:appendix-l3qa-diagnostics}--\ref{sec:appendix-planning-controls}.
Appendix~\ref{sec:appendix-qualitative} shows CDQA, rationale, and CP outputs.
Appendices~\ref{sec:appendix-communication} and~\ref{sec:appendix-scope}
cover transmission cost, release information, and limitations.

\section{Benchmark Construction and Validation}
\label{sec:appendix-annotation-validation}

\paragraph{Scene evidence and annotation generation.} The L1--L4 layers in
Table~\ref{tab:annotation-layers} organize source records into supervision for
CDQA and CP. L1 matches cooperative object labels across the paired views and
combines source attributes with track-derived motion states. Source
assignments identify matched records; an unmatched label alone does not
establish that an object is invisible in the other image.
Qwen3-VL-32B~\citep{bai2025qwen3vl} produces the L2 scene descriptions, L3
CDQA drafts, and L4 rationale drafts from paired images and structured scene
records. Scene descriptions cover the ego view, roadside view, and additional
roadside observations. CDQA uses fixed question templates; computable answers
and their supporting facts are derived from object visibility, geometry, and
motion records, while open-ended answers retain model-generated explanations
after consistency checks.

Rationales are generated for individual planning frames. Their annotation
prompts include the images, scene descriptions, selected object records, and
recorded command and trajectory as references for the observed behavior.
The resulting rationale contains a scene overview, V2X-aware critical objects,
and decision reasoning. Checks remove explicit references to future-target
fields from this text and preserve the source labels of referenced objects.
CDQA answers are not used to generate rationales. The recorded motion supplies
the separate command and waypoint targets, while inference uses the paired
images and task prompt (Section~\ref{sec:method}).

\paragraph{Planning target construction.}
For each observation, subsequent ego positions are transformed into the ego
coordinate frame at that instant and sampled every 0.5 seconds through
3.0 seconds. The resulting six waypoints use meters, with rightward lateral
position and forward longitudinal position positive. Requiring the complete
horizon yields 2,129 examples; the scene-disjoint split contains 1,475 training
and 654 validation examples. A fixed rule on the motion records supplies the
six-class command label used by the structured head.

\paragraph{Manual inspection.}
After the automatic checks, every scene description, CDQA record, and rationale
was inspected against the paired images through a review interface that shows
both views beside the annotation. Reviewers mark each item pass or reject and,
on reject, record one or more issue categories: scene overview, critical
objects and V2X evidence, decision reasoning, action, image grounding, and
other. Rejected items were corrected or regenerated and inspected again, so the
released annotations pass this check. To estimate the first-pass error rate,
two reviewers independently audited the same 100-item sample of CDQA records
and rationales; about 10\% did not pass on first inspection. The pass criterion
is the absence of errors that contradict the paired images or the structured
records, not a graded quality judgment.

Table~\ref{tab:annotation-quality} summarizes the coverage and consistency
checks applied before training, independently of baseline results.

\begin{table}[!htbp]
\caption{\textbf{Annotation validation}. All checks are independent of
model training.}
\label{tab:annotation-quality}
\centering
\small
\setlength{\tabcolsep}{4pt}
\renewcommand{\arraystretch}{1.04}
\begin{tabular}{lr}
\toprule
Check & Result \\
\midrule
\tablegroupgray{2}{Annotation coverage}
Scene descriptions & 2,196 / 2,196 \\
CDQA annotations & 35,136; 16 per frame \\
Three-part rationales & 2,129 / 2,129 \\
CP train / validation examples & 1,475 / 654 \\
\tablegroupgray{2}{Consistency checks}
Schema and field completeness & Pass \\
Direction consistency & Pass \\
Object-source consistency & Pass \\
Structured-target separation & Pass \\
Future-target exclusion & Pass \\
\tablegroupgray{2}{Manual inspection}
Scene descriptions inspected & All \\
CDQA records inspected & All \\
Three-part rationales inspected & All \\
Sampled two-reviewer audit & 100 items; $\sim$10\% corrected \\
\bottomrule
\end{tabular}
\end{table}

\FloatBarrier

\begin{figure}[!htbp]
\centering
\includegraphics[width=\linewidth]{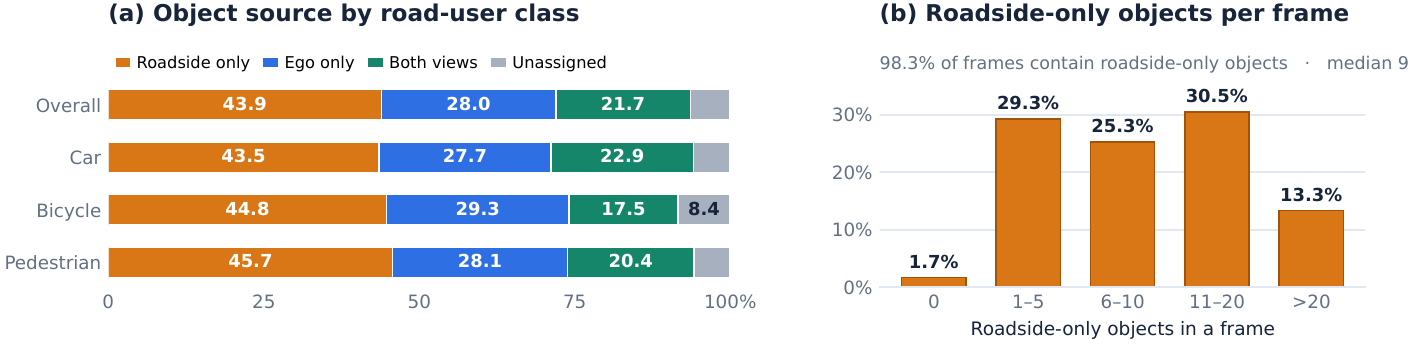}
\caption{\textbf{Roadside evidence coverage in \dataset{}}. (a) Source
assignments for 54,171 annotated objects. (b) Roadside-only objects across 2,196
paired frames. Roadside-only evidence occurs in 98.3\% of frames (median 9
objects). The panels summarize complementary object records available for
annotation; source assignments follow the matching convention in Section~\ref{sec:benchmark}.}
\label{fig:evidence-statistics}
\end{figure}

\paragraph{Object-source distribution.}
Figure~\ref{fig:evidence-statistics} breaks down source assignments by road-user
class and the number of roadside-only objects per frame. These distributions
summarize the complementary object records available for annotation.

\FloatBarrier

\section{Annotation Templates and Model Prompts}
\label{sec:appendix-prompts}

The annotation catalog defines the questions released with the benchmark;
the model prompts specify how CDQA and CP are queried at evaluation time.
The templates below document these distinct uses~\citep{xie2025drivebench}.

\paragraph{Annotation catalog.}
Table~\ref{tab:qa-annotation-catalog} presents the complete 16-question catalog
using generic object placeholders. Each frame stores the instantiated questions,
answers, concise reasons, and, for object-related questions, supporting object
evidence. Evaluation prompts
replace source-specific object identifiers with camera-neutral tokens. The five
blue-marked tasks support objective scoring across roadside-input conditions;
the remaining templates are also released for training or separate analysis.

\begin{table}[!htbp]
\caption{\textbf{Complete CDQA annotation catalog}. All 16 question templates
are instantiated for every paired frame.
A \textcolor{benchBlue}{blue check} marks a closed task used in the reported
CDQA score; a \textcolor{black!45}{gray check} marks a task evaluated as a
source-role diagnostic outside that score. For presentation, \(O_a\), \(O_b\), and \(O_c\) denote generic
object placeholders; \(O_r\) denotes the queried roadside-grounded object.
The maneuver choice question is scored against the recorded ego maneuver
(Appendix~\ref{sec:appendix-scope}).}
\label{tab:qa-annotation-catalog}
\centering
\footnotesize
\setlength{\tabcolsep}{2.8pt}
\renewcommand{\arraystretch}{1.06}
\begin{tabular}{>{\raggedright\arraybackslash}p{0.18\linewidth}
                >{\raggedright\arraybackslash}p{0.45\linewidth}
                >{\raggedright\arraybackslash}p{0.24\linewidth}
                >{\centering\arraybackslash}p{0.04\linewidth}}
\toprule
Template & Instantiated question & Answer form & Use \\
\midrule
\rowcolor{tableBandGray}
\multicolumn{4}{l}{\textbf{Prediction}}\\
Object behavior I & What is the likely behavior of $\langle O_a\rangle$ over the next three seconds? & Continue / Slow or stop / Accelerate / Change lane or turn & \\
Object behavior II & What is the likely behavior of $\langle O_b\rangle$ over the next three seconds? & Continue / Slow or stop / Accelerate / Change lane or turn & \\
\rowcolor{tableFocusBlue}
Relative distance I & Will $\langle O_a\rangle$ be closer, farther, or at roughly the same distance after three seconds? & Closer / Farther / Same / Unable to determine & \textcolor{benchBlue}{\ding{51}} \\
\rowcolor{tableFocusBlue}
Relative distance II & Will $\langle O_b\rangle$ be closer, farther, or at roughly the same distance after three seconds? & Closer / Farther / Same / Unable to determine & \textcolor{benchBlue}{\ding{51}} \\

\rowcolor{tableBandBlue}
\multicolumn{4}{l}{\textbf{Maneuver QA}}\\
\rowcolor{tableFocusBlue}
Maneuver choice & Which maneuver is safest for the ego vehicle? & Maintain / Slow / Brake / Change lane / Accelerate & \textcolor{benchBlue}{\ding{51}} \\
Unsafe maneuvers & Which ego maneuvers would be unsafe because of roadside-only objects? & Open list of maneuver phrases, or None & \\
Attention priority & Rank $\langle O_a,O_b,O_c\rangle$ by relevance to the immediate driving decision. & Ordered list of object tokens & \\
Counterfactual priority & How would the ranking change if $\langle O_r\rangle$ were not visible? & Ordered list with $O_r$ removed & \\

\rowcolor{tableBandGray}
\multicolumn{4}{l}{\textbf{V2X evidence analysis}}\\
Risk without roadside & What is the collision risk involving $\langle O_r\rangle$ without matched roadside evidence? & Low / Medium / High & \\
Risk with roadside & What is the collision risk involving $\langle O_r\rangle$ with matched roadside evidence? & Low / Medium / High & \\
Decision change & Does matched roadside evidence change the best ego decision? & Yes / No & \textcolor{black!45}{\ding{51}} \\
Critical object and change & Which roadside-only object most affects the decision, and how does the decision change? & Object token plus a short change description & \\
Scene evidence tier & How relevant is matched roadside evidence to safe driving in this scene? & Low / Medium / High & \textcolor{black!45}{\ding{51}} \\
Critical roadside object & Which roadside-only object is most important, and what factor makes it important? & Object token plus factor & \\
\rowcolor{tableFocusBlue}
Roadside-only identities & Which neutral object tokens are visible only in the matched roadside view? & List of $O_k$ tokens, or an empty list & \textcolor{benchBlue}{\ding{51}} \\
\rowcolor{tableFocusBlue}
Nearest roadside distance & What is the distance to the nearest object visible only in the matched roadside view? & Numeric distance in meters, or N/A & \textcolor{benchBlue}{\ding{51}} \\
\bottomrule
\end{tabular}
\end{table}

\FloatBarrier

\paragraph{Unscored templates.}
The eleven templates outside the reported score carry the same answer-and-reason
structure as the scored ones. Table~\ref{tab:unscored-cdqa-example} shows the
released annotations for one validation frame, omitting the second
object-behavior question because it repeats the first on another object. The two
collision-risk templates are annotated as a pair on the same object, so the
change from one to the other records what the matched roadside view contributes
to that frame.

\begin{table}[!htbp]
\caption{\textbf{Unscored CDQA annotations for one validation frame}
(\texttt{val\_003446}, ten roadside-only objects). Answers, less their
multiple-choice option letter, and reasons are reproduced verbatim from the
released annotations; \([I\ast]\) and \([E\ast]\) are the released
identifiers for infrastructure- and ego-observed objects, which the
evaluation prompts replace with camera-neutral tokens, and motion labels
such as \texttt{going\_straight} come from the recorded
motion-state vocabulary. These templates are released for training and
separate analysis and are not part of the reported CDQA accuracy.}
\label{tab:unscored-cdqa-example}
\centering
\footnotesize
\setlength{\tabcolsep}{3pt}
\renewcommand{\arraystretch}{1.08}
\begin{tabular}{>{\raggedright\arraybackslash}p{0.17\linewidth}
                >{\raggedright\arraybackslash}p{0.19\linewidth}
                >{\raggedright\arraybackslash}p{0.58\linewidth}}
\toprule
Template & Annotated answer & Annotated reason \\
\midrule
\rowcolor{tableBandGray}
\multicolumn{3}{l}{\textbf{Prediction}}\\
Object behavior I & continue & [I1] is going\_straight with a smooth speed of 14.15 m/s, indicating continuous forward motion. \\
\rowcolor{tableBandGray}
\multicolumn{3}{l}{\textbf{Maneuver QA}}\\
Unsafe maneuvers & change lane/turn; accelerate & Lane-turning or accelerating could bring ego into conflict with [I3] (stopped\_at\_signal) or [I2] (crossing/turning) in rear-left, which are only visible via V2X. \\
Attention priority & I1; I8; I9 & [I1] is closest (37.8m) and in same direction, posing immediate rear-end risk; [I8] and [I9] are farther (122.9m, 124.4m) and less immediate. \\
Counterfactual priority & I8; I9 & Without [I1], [I8] and [I9] are the closest rear objects, both at $\sim$123m, so they become equal priority. \\
\rowcolor{tableBandGray}
\multicolumn{3}{l}{\textbf{V2X evidence analysis}}\\
Risk without roadside & High & [I1] is 37.8m behind and moving at 14.15 m/s in same direction; without V2X, ego may not detect it, risking rear-end collision. \\
Risk with roadside & Low & With V2X, [I1] is visible and ego can maintain safe distance; no immediate threat as speeds are similar and ego is not accelerating. \\
Critical object and change & [I1] supports rear-side monitoring while continuing straight & [I1] is an infra-only bicycle at 37.8m to the rear-left with going straight motion. Its main V2X value is rear-side awareness for blind-side clearance and lane keeping, not a direct forward-path obstacle. V2X helps maintain clearance while the ego continues straight and keep the action aligned with the observed corridor. \\
\bottomrule
\end{tabular}

\end{table}

\FloatBarrier

\subsection{CDQA evaluation prompts}

Prompts use each backbone's native chat template as user messages, without an
additional system message. Figure~\ref{fig:appendix-prompt-templates} summarizes
the input and response requirements.

\begin{figure}[!htbp]
\centering
\begin{tcolorbox}[
  enhanced,
  colback=white,
  colframe=bothGreen!72!black,
  colbacktitle=bothGreen!10,
  coltitle=bothGreen!50!black,
  fonttitle=\bfseries,
  title={CDQA: inference prompt under roadside-input conditions},
  boxrule=0.6pt,
  arc=2pt,
  left=7pt,right=7pt,top=5pt,bottom=5pt
]
\small
\renewcommand{\arraystretch}{1.16}
\begin{tabular}{@{}>{\bfseries\color{bothGreen!50!black}}p{0.17\linewidth}p{0.76\linewidth}@{}}
Condition & \textbf{Roadside input}: valid, ego-only, blank, or shuffled. The ego image and scene target remain fixed. \\
Catalog & Each neutral token $O_k$ has a category, current position, and distance. Token names do not reveal which camera observed the object. \\
Grounding rule & Answer from the supplied images and current catalog only. Infer roadside-only visibility by comparing the two annotated views; treat missing evidence as unavailable. \\
Task bundle & Answer five objective questions and two source-role diagnostics for the same frame and condition. \\
Response rule & Return answer-only structured output with one value per task. Use a list for roadside-only identities and a numeric distance or ``N/A'' for the nearest-distance question. Do not add explanations, markdown, or extra fields. \\
\end{tabular}

\vspace{4pt}
\setlength{\tabcolsep}{5pt}
\renewcommand{\arraystretch}{1.12}
\begin{tabular}{@{}p{0.465\linewidth}p{0.465\linewidth}@{}}
\cellcolor{tableGoodGreen}\textbf{Valid}\quad Ego image and paired roadside image; retain the current-scene catalog. &
\cellcolor{tableRefBlue}\textbf{Ego-only}\quad Ego image only; withhold current-scene roadside evidence. \\
\cellcolor{tableNeutralGray}\textbf{Blank}\quad Ego image and a black roadside frame; keep the scene target unchanged. &
\cellcolor{infraOrange!12}\textbf{Shuffled}\quad Ego image and a scene-disjoint roadside image; donor objects are not matched evidence. \\
\end{tabular}

\vspace{4pt}
\begin{tcolorbox}[
  colback=tableNeutralGray,
  colframe=tableVariantGray,
  boxrule=0.35pt,
  arc=1pt,
  left=5pt,right=5pt,top=3pt,bottom=3pt
]
\footnotesize
\textbf{Objective fields:} ego-view prediction $\cdot$ roadside-view prediction
$\cdot$ maneuver QA $\cdot$ roadside-only objects $\cdot$ nearest roadside distance

\textbf{Diagnostic fields:} roadside constraint $\cdot$ evidence tier
\end{tcolorbox}
\end{tcolorbox}
\caption{\textbf{CDQA evaluation prompt overview}. Scene content and
object catalogs are instantiated per sample. Field names and instructions are
summarized for readability; Table~\ref{tab:grounded-qa-templates} gives the
question templates. The five objective fields define Overall accuracy, while
the two source-role fields describe responses to roadside evidence.}
\label{fig:appendix-prompt-templates}
\end{figure}

\paragraph{CDQA evaluation.}
Every model receives the same seven-question frame--condition bundle. Decoding
is deterministic (sampling disabled, repetition penalty 1.0, and at most 256
new tokens), and the image budget is 1,048,576 pixels. Five tasks admit
objective exact-answer scoring; two describe the response to roadside evidence
and are excluded from Overall accuracy. One record-derived template is
released but not evaluated: because object tokens are ordered by distance,
both its object and its factor follow from the nearest-roadside-distance task
that is already scored. Table~\ref{tab:grounded-qa-templates} gives the
complete inference template and eligible population for each task. No
evaluator model or judge prompt is used.

\begin{table}[!htbp]
\caption{\textbf{CDQA inference templates}. Wording and options are
reproduced from the inference prompt, in which ``infrastructure view'' denotes
the roadside view; items are counted per input condition. Blue rows are the
five objective tasks included in Overall accuracy; gray rows are response
diagnostics excluded from that average. $[O_e]$ and $[O_r]$ denote the queried
neutral token grounded in the ego and roadside views. The maneuver QA
reference answer is the recorded ego maneuver, not an independent safety
judgment (Appendix~\ref{sec:appendix-scope}).}
\label{tab:grounded-qa-templates}
\centering
\footnotesize
\setlength{\tabcolsep}{3.0pt}
\renewcommand{\arraystretch}{1.05}
\begin{tabular}{>{\raggedright\arraybackslash}p{0.15\linewidth}
                >{\raggedright\arraybackslash}p{0.40\linewidth}
                >{\raggedright\arraybackslash}p{0.32\linewidth}
                >{\raggedleft\arraybackslash}p{0.05\linewidth}}
\toprule
Task & Question template & Answer options & Items \\
\midrule
\rowcolor{tableFocusBlue}
Ego-view prediction & Will $[O_e]$ be closer/farther in 3s? & A. Closer / B. Farther / C. Roughly the same / D. Unable to determine & 412 \\
\rowcolor{tableFocusBlue}
Roadside-view prediction & Will $[O_r]$ be closer/farther in 3s? & A. Closer / B. Farther / C. Roughly the same / D. Unable to determine & 487 \\
\rowcolor{tableFocusBlue}
Maneuver QA & Safest ego action? & A. Maintain current speed / B. Slow down / C. Brake firmly / D. Change lane / E. Accelerate to clear intersection & 654 \\
\rowcolor{tableFocusBlue}
Roadside-only objects & Comparing the two annotated views, which neutral object tokens are visible only in the matched infrastructure view? & JSON list of $O_k$ tokens, or an empty list if none are supported & 654 \\
\rowcolor{tableFocusBlue}
Nearest roadside distance & First compare the annotated views to identify infrastructure-only $O_k$ tokens, then use their current catalog distances: what is the nearest such distance? & Short numeric string such as ``7.3 m'', or N/A if none are valid & 654 \\
\midrule
\rowcolor{tableNeutralGray}
Roadside constraint & Does matched infrastructure evidence reveal an additional object that constrains the ego's safe maneuver in this scene? & Yes / No & 654 \\
\rowcolor{tableNeutralGray}
Evidence tier & What is the matched-infrastructure evidence tier for the ego's safe maneuver? & Low / Medium / High, each defined in the prompt & 654 \\
\bottomrule
\end{tabular}
\end{table}

\FloatBarrier

\subsection{CP prompts and rationale format}
\label{sec:appendix-cot-template}

The three-part CoT is a text output of \model{} and an auxiliary supervision
target during training. Figure~\ref{fig:appendix-cot-template} gives the full
prompt for the standard CP configuration and the corresponding response
format. The three sections are the scene overview, the V2X-aware critical objects, and
the decision reasoning. Command classes and numerical waypoints
remain separate structured outputs.

\begin{center}
\begin{minipage}{\linewidth}
    \centering
    \includegraphics[width=\linewidth]{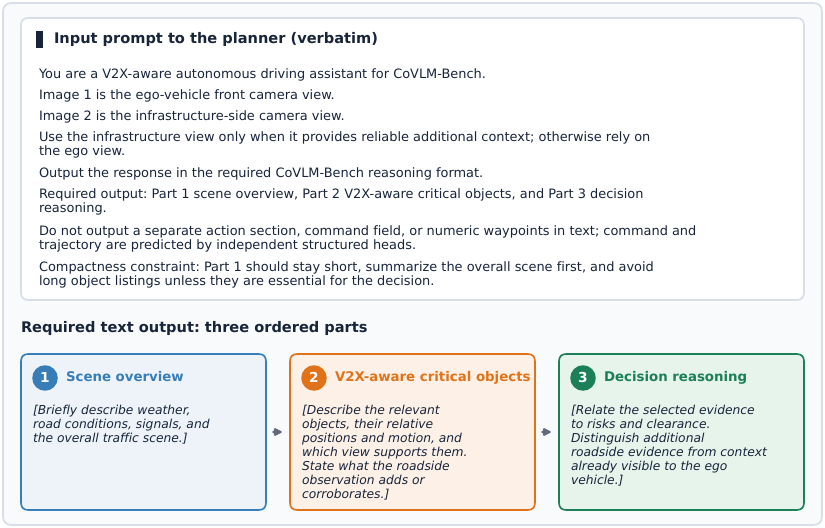}
\captionsetup{type=figure,hypcap=false}
\caption{\textbf{CoT prompt and response format}. The upper card reproduces
the input prompt of the standard CP configuration. The three response headings follow the
training-target format; italic bracketed text describes the content of each
part rather than fixed response wording.}
\label{fig:appendix-cot-template}
\end{minipage}
\end{center}

\paragraph{Zero-shot CP prompt.}
Zero-shot VLMs receive the ego image followed by the matched roadside image.
The prompt requests a command from GO\_STRAIGHT, TURN\_LEFT, TURN\_RIGHT,
LATERAL\_SHIFT, STOP, or SLOW\_DOWN, together with six waypoints at
0.5-second intervals through 3.0 seconds. It defines coordinates in meters
relative to the ego vehicle, with lateral position positive to the right and
longitudinal position positive forward. The response contains only a
\texttt{command} field and a \texttt{waypoints} field, encoded as JSON.
Decoding is deterministic, uses at most 256 new tokens, and caps each image
at 262,144 pixels. This text-generation evaluation uses the unadapted VLMs
listed in Table~\ref{tab:external-baselines}.

\FloatBarrier

\section{Training and Evaluation Protocols}
\label{sec:appendix-reproducibility-release}

\paragraph{Task configurations.}
The CDQA-adapted configuration of \model{} is fine-tuned on single-question
CDQA records and evaluated with the same seven-question bundle as all other
VLMs. It uses the same backbone, LoRA settings, and image cap as the CP
configuration, with two epochs, learning rate \(2\times10^{-5}\), batch size
1, accumulation 8, seed 20260524, and at most 192 target tokens, over all
sixteen templates on the training split. The CP-adapted configuration uses the
planning objectives in Section~\ref{sec:training-objective-interface}. Its
standard CP configuration adds three-part rationale supervision without QA
initialization. These are task-specific adaptations of the same framework; the
supervision study in Appendix~\ref{sec:appendix-supervision-ablations} also
initializes CP training from the CDQA-adapted checkpoint.

\paragraph{CP training and runtime.}
The standard CP configuration uses Qwen3-VL-8B in bf16, LoRA rank/alpha/dropout
16/32/0.05, gradient checkpointing, mean-pooled structured heads, and images
capped at 262,144 pixels. Ten-epoch training uses seed 20260524, batch size 1,
accumulation 8, learning rate \(5\times10^{-5}\), weight decay 0.01, warmup
ratio 0.03, cosine scheduling, and loss weights
\(\lambda_{\mathrm{lm}}=1.0\), \(\lambda_{\mathrm{cmd}}=1.0\),
\(\lambda_{\mathrm{wp}}=2.0\), and \(\lambda_{\mathrm{fde}}=1.0\).
Command classes are not reweighted. Runtime profiling uses batch size
one on an RTX~4090, with
20 warmup examples and the same ordered 100-example subset for every method.

\paragraph{Backbone comparison.}
The alternative backbones use the same training split, rationale targets,
LoRA settings, loss weights, and ten-epoch schedule. Evaluation is performed
every two epochs. Checkpoint selection minimizes validation FDE among models
predicting more than one command class, with balanced command accuracy used
to break ties.
The ego-status variant additionally receives motion features derived from
current and preceding ego poses. The standard CP configuration uses images and the
task prompt only. Runtime measurements cover the structured planning forward
pass, excluding rationale generation, and report peak allocated GPU memory.

\subsection{Metrics and scoring}
\label{sec:appendix-metric-notes}
This section defines the trajectory, command, and CDQA metrics used by
the main tables and appendix diagnostics.

\paragraph{Trajectory metrics.}
All planning rows use ego-relative waypoints on the same 654-frame validation
split. For validation sample \(i\), time \(t\), predicted waypoint
\(\hat{\mathbf{p}}_{i,t}\), and reference waypoint \(\mathbf{p}_{i,t}\), the
horizon-wise displacement error is
\begin{equation}
\mathrm{L2}(t) =
\frac{1}{N}\sum_{i=1}^{N}
\left\lVert \hat{\mathbf{p}}_{i,t}-\mathbf{p}_{i,t}\right\rVert_2,
\qquad
\mathrm{FDE}=\mathrm{L2}(3.0\,\mathrm{s}),
\end{equation}
where \(N=654\) for the validation protocol.
Tables~\ref{tab:external-baselines} and~\ref{tab:input-control-planning}
report L2 at the first five horizons and FDE at the final horizon; lower is
better.

\paragraph{Command metrics.}
Following the high-level maneuver reporting used in V2X planning baselines
\citep{yu2025univ2x,song2025unimmv2x}, we report command accuracy as a
secondary command metric.
Command accuracy measures whether the predicted high-level maneuver matches the
released command label,
\begin{equation}
\mathrm{Cmd} =
\frac{1}{N}\sum_{i=1}^{N}
\mathbf{1}\!\left[\hat{c}_i=c_i\right],
\end{equation}
where \(c_i\) and \(\hat{c}_i\) are the reference and predicted commands.
Balanced command accuracy (Bal. Acc.) averages class-wise accuracy over the
command classes present in the validation split,
\begin{equation}
\mathrm{BalAcc} =
\frac{1}{|\mathcal{C}_{\mathrm{val}}|}
\sum_{c\in\mathcal{C}_{\mathrm{val}}}
\frac{\sum_{i=1}^{N}\mathbf{1}[c_i=c]\mathbf{1}[\hat{c}_i=c_i]}
{\sum_{i=1}^{N}\mathbf{1}[c_i=c]}.
\end{equation}
Here \(\mathcal{C}_{\mathrm{val}}\) is the set of command classes present in
the validation split. Command accuracy weights examples equally, whereas
balanced accuracy weights these classes equally. This distinction matters for
the unequal command frequencies in the benchmark.

\paragraph{Zero-shot output scoring.}
Parse (\%) is the fraction of all 654 responses from which the evaluator
extracts a supported command and six finite two-dimensional waypoints. This
measures output-format compliance. An unparseable response receives an unknown
command and six zero-displacement waypoints; it remains in every metric's
denominator. Parsed coordinates are scored as generated, without clipping to
a speed or distance threshold. These rules give all zero-shot models the same
evaluation population and prevent format failures from being discarded.

\paragraph{CDQA accuracy.}
CDQA prompts use camera-neutral object identifiers and require one closed answer
for each objective task. We report exact-answer accuracy for ego-view
prediction, roadside-view prediction, maneuver QA, roadside-only object
identity, and nearest roadside distance. A frame--task pair is invalid under the CDQA
construction rules when the future observation is missing or the queried object
is ungrounded; the same 409 pairs are
excluded under every condition. The remaining task populations contain 412,
487, 654, 654, and 654 items, respectively, for 2,861 eligible items per
condition. Overall accuracy is the micro-average over those items. Best absent
is the higher of ego-only and blank-roadside accuracy. Shuffled roadside input
tests scene mismatch and is not included in that absent-view control.

Maneuver QA selects among maintain current speed, slow down, brake firmly,
change lane, and accelerate to clear intersection.
Command prediction uses the separate GO\_STRAIGHT, TURN\_LEFT, TURN\_RIGHT,
LATERAL\_SHIFT, STOP, and SLOW\_DOWN labels. Their targets and answer spaces
differ, so maneuver QA accuracy and command accuracy are reported separately.

The maneuver QA reference option is selected by a fixed rule on the recorded
future ego trajectory, using the longitudinal displacement at the end of the
horizon, the lateral displacement, and the speed difference between the first
and last steps. The task therefore scores recovery of the maneuver the ego
vehicle executed. Appendix~\ref{sec:appendix-l3qa-diagnostics} reports its
response to input controls, and Appendix~\ref{sec:appendix-scope} explains the
distinction between recorded-maneuver agreement and the safety-oriented prompt
wording.

\subsection{External baseline alignment}
\label{sec:appendix-baselines-qualitative}
\paragraph{External-baseline conversion.}
External baselines are reported only when their outputs can be mapped onto the
shared validation protocol. UniV2X~\citep{yu2025univ2x} has 675 native rows and
overlaps the \dataset{} validation protocol on 654 rows; UniMM-V2X~\citep{song2025unimmv2x}
is exported on the same 654-frame split with compatible command metrics.
MAP~\citep{yin2025map} provides 675 native outputs and 654 aligned examples,
which are scored against the \dataset{} waypoint reference. UniV2X
and MAP provide outputs for every aligned example. For UniV2X, native waypoint
coordinates are converted from forward-lateral form to the \dataset{}
ego-relative coordinate convention as
\begin{equation}
\left[x_{\mathrm{lateral}},\,y_{\mathrm{forward}}\right]
=
[-l,\,f],
\end{equation}
where \(f\) is the native forward coordinate and \(l\) is the native lateral
coordinate. For MAP, saved waypoints are converted as
\begin{equation}
\left[x_{\mathrm{lateral}},\,y_{\mathrm{forward}}\right]
=
[-y_{\mathrm{saved}},\,-x_{\mathrm{saved}}],
\end{equation}
where \(x_{\mathrm{saved}}\) and \(y_{\mathrm{saved}}\) denote MAP's saved
trajectory coordinates. FDE is computed only after coordinate conversion and
six-step horizon alignment, which places Table~\ref{tab:external-baselines} on a
common output evaluation.

\paragraph{Native UniV2X protocol.}
UniV2X~\citep{yu2025univ2x} was originally evaluated on its own native
DAIR-V2X-Seq / V2X-Seq-SPD protocol, which differs from \dataset{} in frame
coverage, trajectory format, and reported task set.
Table~\ref{tab:external-baselines} therefore reports it under the shared
654-frame six-step evaluation described above. Cmd Acc. and Bal. Acc. are
reported only when a method exposes a
compatible command prediction or command head.

\FloatBarrier

\section{Supplementary CDQA Results}
\label{sec:appendix-l3qa-diagnostics}

Table~\ref{tab:appendix-cdqa-conditions} expands the main CDQA comparison into
four input conditions and three task groups: prediction, maneuver QA, and V2X
evidence. It uses the same model roster and 2,861 eligible items per condition.
Prediction pools 899 items, maneuver QA has 654, and V2X evidence pools 1,308.
Best absent in the main table selects the stronger of ego-only and blank;
shuffled tests whether the roadside image matches the queried scene.

The V2X group separates the conditions most clearly. A mismatched roadside image
reduces V2X accuracy to the level reached without any roadside view: 62.4\% to
1.8\% for Qwen3-VL-32B and 39.5\% to 2.7\% for CDQA-adapted \model{}, against
0.9\% and 1.3\% for ego-only input. Answering these questions thus requires the
roadside image of the same scene. Maneuver QA shows a different pattern:
several models score higher with ego-only input, while both adapted
configurations vary little across conditions. Its reference labels describe
recorded ego motion (Appendix~\ref{sec:appendix-metric-notes}).

The last block of Table~\ref{tab:appendix-cdqa-conditions} pools the four
remaining objective tasks, which is Overall with the 654 maneuver QA items
removed, isolating the four object-related tasks. Over these 2,207 items,
every model has a larger advantage over its best absent-view control: the gap
increases from \(+33.2\) to \(+46.8\) points for Qwen3-VL-32B and from
\(+25.3\) to \(+32.3\) for CDQA-adapted \model{}; LLaVA-OneVision-7B has a
positive gap of \(+0.6\). The main Overall metric continues to cover all five
objective tasks.

\begingroup
\small
\setlength{\tabcolsep}{6.0pt}
\renewcommand{\arraystretch}{1.06}
\begin{longtable}{lrrrr}
\caption{\textbf{CDQA accuracy by roadside-input condition} (\%). All models use the same 2,861 eligible items. Prediction pools 899 items, maneuver QA has 654, and V2X evidence pools 1,308. The last block excludes maneuver QA, leaving 2,207 items. Adapted rows use task-specific checkpoints of \model{} with Qwen3-VL-8B.}
\label{tab:appendix-cdqa-conditions}\\
\toprule
Model & Valid & Ego-only & Blank & Shuffled \\
\endfirsthead
\multicolumn{5}{l}{\small Table \thetable{} continued: CDQA input conditions}\\
\toprule
Model & Valid & Ego-only & Blank & Shuffled \\
\endhead
\midrule
\multicolumn{5}{r}{\small Continued on next page}\\
\endfoot
\bottomrule
\endlastfoot
\midrule
\rowcolor{tableBandGray}\multicolumn{5}{l}{\textbf{Overall}}\\*
Qwen3-VL-4B & 20.4 & 14.2 & 13.8 & 13.7 \\*
Qwen3-VL-8B & 24.7 & 19.5 & 19.0 & 19.6 \\*
Qwen3-VL-32B & 51.1 & 17.0 & 17.8 & 17.5 \\*
Qwen2.5-VL-3B & 24.9 & 19.3 & 19.0 & 22.4 \\*
Qwen2.5-VL-7B & 19.4 & 12.5 & 11.5 & 13.1 \\*
InternVL3-8B & 29.8 & 19.0 & 18.3 & 19.2 \\*
LLaVA-OneVision-7B & 23.4 & 23.8 & 24.4 & 21.6 \\*
Idefics3-8B & 16.9 & 9.8 & 0.0 & 4.1 \\*
\model{} (CP-adapted) & 25.8 & 20.8 & 19.4 & 21.1 \\*
\model{} (CDQA-adapted) & 39.6 & 14.2 & 14.1 & 15.7 \\
\midrule
\rowcolor{tableBandGray}\multicolumn{5}{l}{\textbf{Prediction}}\\*
Qwen3-VL-4B & 48.3 & 18.1 & 19.2 & 23.9 \\*
Qwen3-VL-8B & 40.7 & 20.1 & 19.0 & 20.2 \\*
Qwen3-VL-32B & 43.5 & 17.2 & 18.2 & 16.2 \\*
Qwen2.5-VL-3B & 42.3 & 18.5 & 17.5 & 31.7 \\*
Qwen2.5-VL-7B & 5.7 & 3.7 & 3.6 & 3.6 \\*
InternVL3-8B & 30.5 & 14.1 & 13.9 & 15.6 \\*
LLaVA-OneVision-7B & 36.3 & 44.3 & 29.3 & 38.2 \\*
Idefics3-8B & 19.4 & 7.6 & 0.0 & 4.9 \\*
\model{} (CP-adapted) & 40.7 & 24.4 & 20.8 & 26.0 \\*
\model{} (CDQA-adapted) & 25.7 & 1.8 & 1.4 & 5.1 \\
\midrule
\rowcolor{tableBandGray}\multicolumn{5}{l}{\textbf{Maneuver QA}}\\*
Qwen3-VL-4B & 20.0 & 35.8 & 32.3 & 25.4 \\*
Qwen3-VL-8B & 46.0 & 56.1 & 55.7 & 56.3 \\*
Qwen3-VL-32B & 38.8 & 49.1 & 51.2 & 50.8 \\*
Qwen2.5-VL-3B & 44.0 & 57.6 & 57.8 & 52.1 \\*
Qwen2.5-VL-7B & 48.5 & 48.2 & 43.9 & 50.2 \\*
InternVL3-8B & 57.5 & 57.8 & 57.6 & 57.5 \\*
LLaVA-OneVision-7B & 33.8 & 37.8 & 58.0 & 36.2 \\*
Idefics3-8B & 29.7 & 28.4 & 0.0 & 10.1 \\*
\model{} (CP-adapted) & 55.5 & 55.8 & 54.7 & 55.2 \\*
\model{} (CDQA-adapted) & 58.9 & 57.2 & 56.9 & 56.3 \\
\midrule
\rowcolor{tableBandGray}\multicolumn{5}{l}{\textbf{V2X evidence}}\\*
Qwen3-VL-4B & 1.3 & 0.8 & 0.8 & 0.8 \\*
Qwen3-VL-8B & 3.0 & 0.8 & 0.8 & 0.8 \\*
Qwen3-VL-32B & 62.4 & 0.9 & 0.8 & 1.8 \\*
Qwen2.5-VL-3B & 3.4 & 0.8 & 0.8 & 1.0 \\*
Qwen2.5-VL-7B & 14.4 & 0.8 & 0.8 & 1.2 \\*
InternVL3-8B & 15.6 & 2.9 & 1.6 & 2.4 \\*
LLaVA-OneVision-7B & 9.4 & 2.8 & 4.3 & 3.0 \\*
Idefics3-8B & 8.8 & 2.1 & 0.0 & 0.5 \\*
\model{} (CP-adapted) & 0.8 & 0.8 & 0.8 & 0.8 \\*
\model{} (CDQA-adapted) & 39.5 & 1.3 & 1.4 & 2.7 \\
\midrule
\rowcolor{tableBandGray}\multicolumn{5}{l}{\textbf{Objective tasks excluding maneuver QA}}\\*
Qwen3-VL-4B & 20.5 & 7.8 & 8.3 & 10.2 \\*
Qwen3-VL-8B & 18.4 & 8.7 & 8.2 & 8.7 \\*
Qwen3-VL-32B & 54.7 & 7.5 & 7.9 & 7.7 \\*
Qwen2.5-VL-3B & 19.3 & 8.0 & 7.6 & 13.5 \\*
Qwen2.5-VL-7B & 10.9 & 1.9 & 1.9 & 2.2 \\*
InternVL3-8B & 21.6 & 7.5 & 6.6 & 7.8 \\*
LLaVA-OneVision-7B & 20.3 & 19.7 & 14.5 & 17.3 \\*
Idefics3-8B & 13.1 & 4.3 & 0.0 & 2.3 \\*
\model{} (CP-adapted) & 17.0 & 10.4 & 8.9 & 11.1 \\*
\model{} (CDQA-adapted) & 33.9 & 1.5 & 1.4 & 3.7 \\
\end{longtable}
\endgroup

\FloatBarrier

\section{Detailed Supervision Ablations}
\label{sec:appendix-supervision-ablations}

\paragraph{QA initialization and rationale supervision.}
Table~\ref{tab:supervision-ablation} expands the main-text comparison in
Table~\ref{tab:supervision-summary} with per-horizon and command metrics. It
separates QA initialization from three-part rationale supervision in
Qwen3-VL-8B. QA initialization starts from the CDQA-adapted checkpoint before
CP training. All four variants use paired images, the same structured heads,
planning targets, and CP training budget, without ego-status input. Rationale
supervision adds the three-part text loss; removing it leaves the command and
trajectory objectives unchanged. The factorial comparison measures its effect
with and without QA initialization, which adds a prior CDQA adaptation stage
while holding the subsequent CP budget fixed. CDQA accuracy after CP training
measures the resulting model's QA performance separately from its planning
accuracy. Against planning targets alone (5.197~m), QA initialization lowers
FDE to 5.083~m and rationale supervision lowers it to 4.911~m. Using both
gives 5.095~m: the two act as substitutes rather than complements, and the
standard CP configuration keeps rationale supervision alone. QA initialization
raises CDQA accuracy from 24.8\% and 25.8\% without it to 37.4\% and 34.9\%
with it.

\begin{table}[!htbp]
\caption{\textbf{QA initialization and rationale supervision}. CP uses the same 654
examples as Table~\ref{tab:external-baselines}; L2 and FDE are in meters
(lower is better). Acc. and Bal. are command accuracy and balanced accuracy.
CDQA reports accuracy (\%) on 2,861 eligible items with the matched roadside
view. The standard CP configuration is highlighted in \blueword{}.}
\label{tab:supervision-ablation}
\centering
\footnotesize
\setlength{\tabcolsep}{3pt}
\renewcommand{\arraystretch}{1.12}
\begin{adjustbox}{max width=\linewidth}
\begin{tabular}{ccrrrrrrrrr}
\toprule
\multicolumn{2}{c}{Supervision}
& \multicolumn{6}{c}{Trajectory L2 (m) $\downarrow$}
& \multicolumn{2}{c}{Command $\uparrow$} & CDQA $\uparrow$ \\
\cmidrule(lr){1-2}\cmidrule(lr){3-8}\cmidrule(lr){9-10}\cmidrule(l){11-11}
\makecell{QA\\init.} & Rationale
& 0.5 s & 1.0 s & 1.5 s & 2.0 s & 2.5 s & FDE & Acc. & Bal. & Acc. (\%) \\
\midrule
No & No & 0.822 & 1.570 & 2.411 & 3.287 & 4.229 & 5.197
& 0.7752 & 0.6221 & 24.8 \\
\tablefocus
No & Yes & 0.812 & 1.500 & 2.283 & 3.119 & 3.998 & 4.911
& 0.7706 & 0.5661 & 25.8 \\
Yes & No & 0.824 & 1.524 & 2.335 & 3.218 & 4.121 & 5.083
& 0.7844 & 0.6172 & 37.4 \\
Yes & Yes & 0.805 & 1.507 & 2.337 & 3.211 & 4.129 & 5.095
& 0.7569 & 0.5467 & 34.9 \\
\bottomrule
\end{tabular}
\end{adjustbox}
\end{table}

\FloatBarrier

\section{Roadside-Input Analysis}
\label{sec:appendix-planning-controls}

Each control preserves the sample, ego image, and planning target. The shuffled
condition assigns a roadside image from another scene in the same split,
without reusing donor images.

\begin{figure}[!htbp]
\centering
\includegraphics[width=\linewidth]{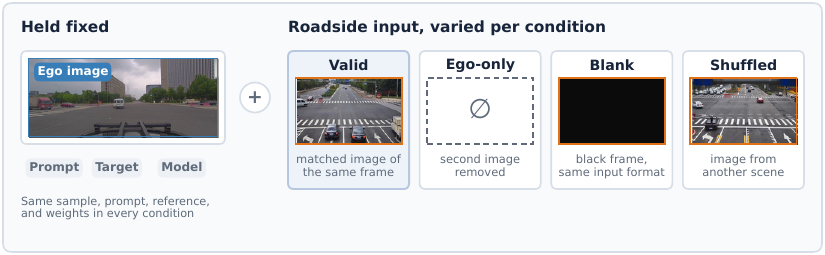}
\caption{\textbf{Roadside-input controls}. The sample, ego image, prompt,
target, and model remain fixed. Valid uses the synchronized roadside image;
ego-only removes the second image; blank supplies a black frame in the same
input format; and shuffled supplies the roadside image of another scene.
Ego-only and blank test roadside availability, whereas shuffled tests scene
correspondence.}
\label{fig:matched-controls}
\end{figure}
\FloatBarrier

On the 513 examples whose rationale annotation lists a roadside-only object
among its critical objects, valid input gives 4.922~m FDE. Ego-only, blank,
and shuffled input give 5.878, 5.191, and 4.972~m, corresponding to
control-minus-valid differences of \(+0.956\), \(+0.269\), and \(+0.051\)~m.
The displayed FDE values are rounded independently; differences are computed
before rounding. Positive differences favor valid input. A fixed rule over
distance and motion produces this critical-object list, so membership reflects
that rule rather than a measured effect on the decision.

Averaged differences summarize the full 654 examples unevenly, so we also
report how often each control is worse than valid input. On the 654 examples,
valid input gives the lower FDE for 63.5\% against ego-only, 52.6\% against
blank, and 51.4\% against shuffled. The shuffled comparison is close to an
even split per example, while the aggregate trajectory and command metrics
favor valid over shuffled input. To describe output sensitivity, we compute
the Euclidean distance between valid and shuffled predictions at each of the
six waypoints and average over time. This mean displacement exceeds 0.1~m on
95.3\% of examples and averages 0.73~m over all 654 examples. None of the
paired predictions is exactly identical. These shifts measure sensitivity to
the supplied view; the FDE comparisons above measure prediction quality.
Table~\ref{tab:appendix-cdqa-conditions} gives the corresponding CDQA
accuracies on the questions that require the matched view.

\FloatBarrier

\clearpage
\section{Qualitative Examples}
\label{sec:appendix-qualitative}

\subsection{CDQA responses}
Figures~\ref{fig:appendix-qa-qualitative}
and~\ref{fig:appendix-qa-qualitative-b} show CDQA responses, and
Figure~\ref{fig:appendix-waypoint-qualitative} shows structured plans, from
paired observations.

\begin{center}
\begin{minipage}{\linewidth}
    \centering
    \includegraphics[width=0.94\linewidth]{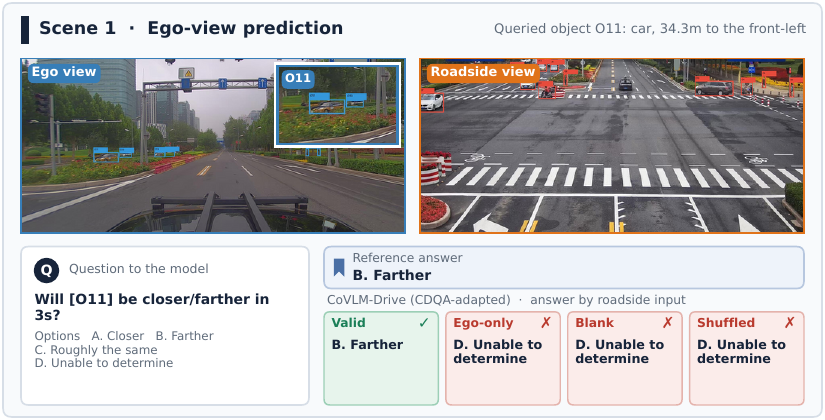}\\[1pt]
    \includegraphics[width=0.94\linewidth]{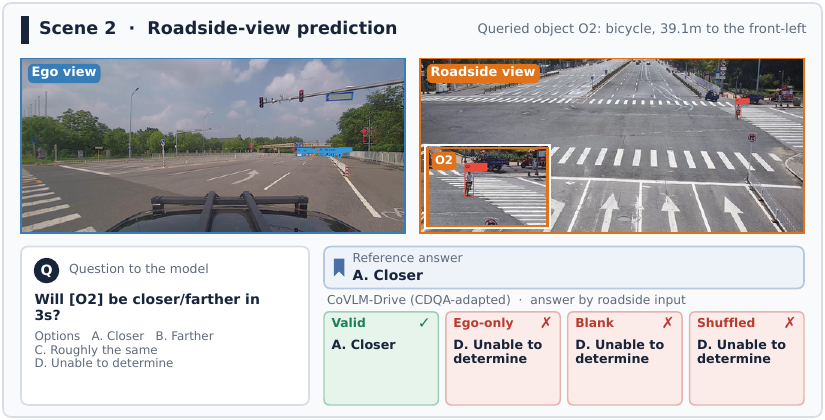}
\captionsetup{type=figure,hypcap=false}
\caption{\textbf{CDQA examples (scenes 1--2)}. Each scene shows the paired
ego and roadside images, the question as posed to the model, the
record-derived reference answer, and the answers of CDQA-adapted \model{}
under the four roadside-input conditions; \cmark{}/\xmark{} mark agreement
with the reference, and insets enlarge the queried object. Scenes~1 and~2
query the future distance of an ego-view and a roadside-view object.}
\label{fig:appendix-qa-qualitative}
\end{minipage}
\end{center}

\clearpage

\begin{center}
\begin{minipage}{\linewidth}
    \centering
    \includegraphics[width=0.94\linewidth]{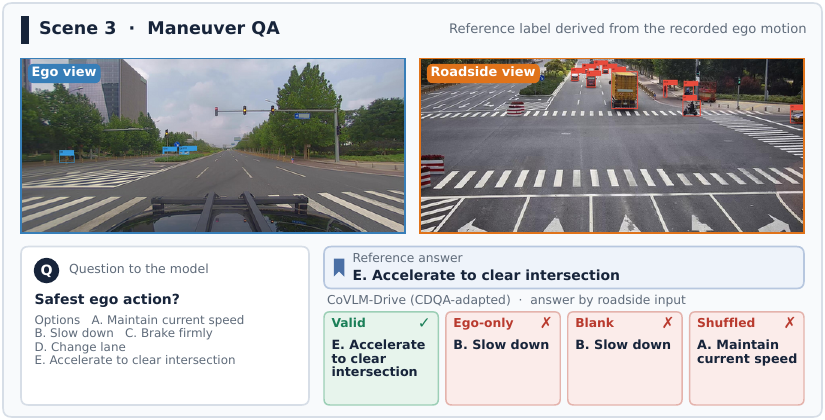}\\[1pt]
    \includegraphics[width=0.97\linewidth]{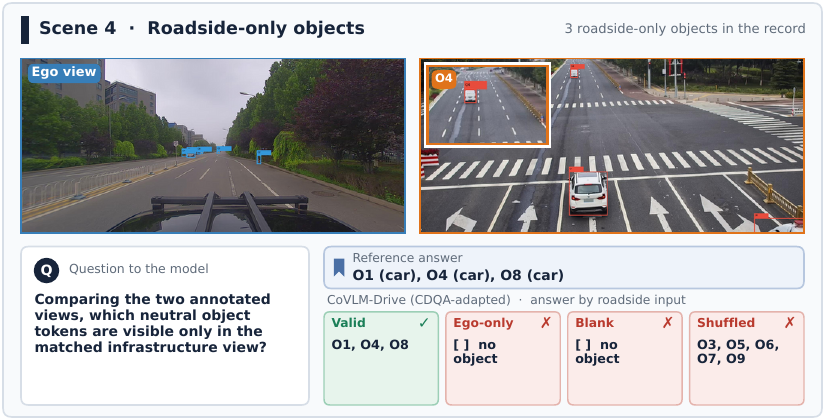}\\[3pt]
    \includegraphics[width=0.97\linewidth]{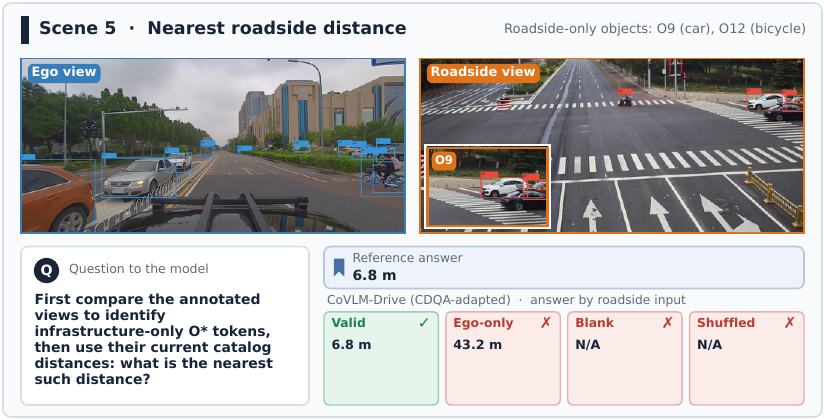}
\captionsetup{type=figure,hypcap=false}
\caption{\textbf{CDQA examples (scenes 3--5)}. Scene~3 compares answers with
the recorded ego maneuver. Scene~4 asks which objects are visible only in the
roadside view and Scene~5 the catalog distance of the nearest such object.
With the matched roadside image both agree with the reference. Withholding or
blanking it empties the list, or makes the distance unavailable or wrong, and
the scene-disjoint shuffled image leads the model to list ego-view objects in
Scene~4.}
\label{fig:appendix-qa-qualitative-b}
\end{minipage}
\end{center}
\clearpage
\subsection{Generated rationales}
\label{sec:appendix-cot-examples}
Two scenes illustrate the three-part text output of the standard CP configuration.

\begin{center}
\begin{minipage}{\linewidth}
    \centering
    \includegraphics[width=\linewidth]{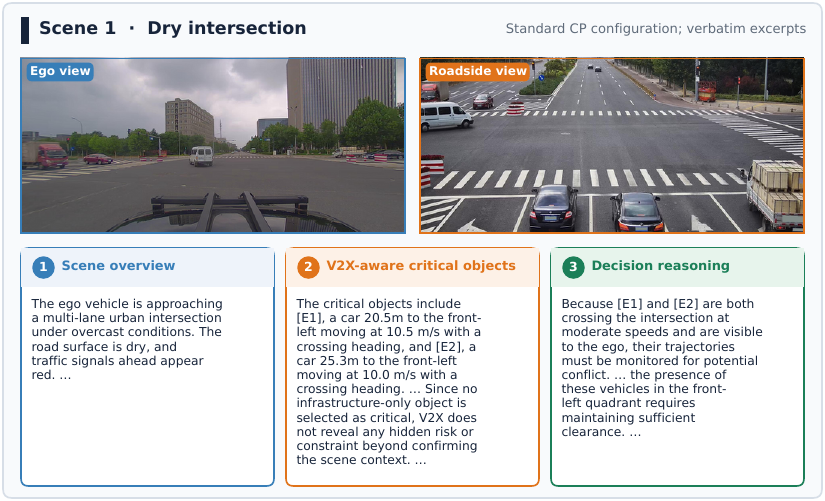}\\[4pt]
    \includegraphics[width=\linewidth]{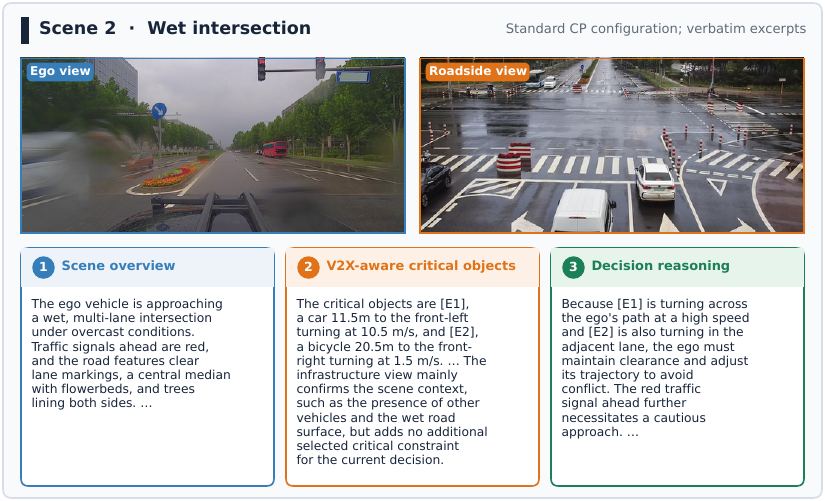}
\captionsetup{type=figure,hypcap=false}
\caption{\textbf{Generated CoT excerpts}. Two validation scenes with verbatim
excerpts from the three response parts generated by the standard CP configuration;
ellipses mark omitted text. [E1] and [E2] are the object identifiers used in
rationale supervision, not the camera-neutral CDQA tokens. The examples
illustrate the three-part rationale format; command and trajectory
predictions are produced by the separate structured heads.}
\label{fig:appendix-cot-qualitative}
\end{minipage}
\end{center}
\clearpage
\subsection{Cooperative planning}
\begin{center}
\begin{minipage}{\linewidth}
    \centering
    \includegraphics[width=0.96\linewidth]{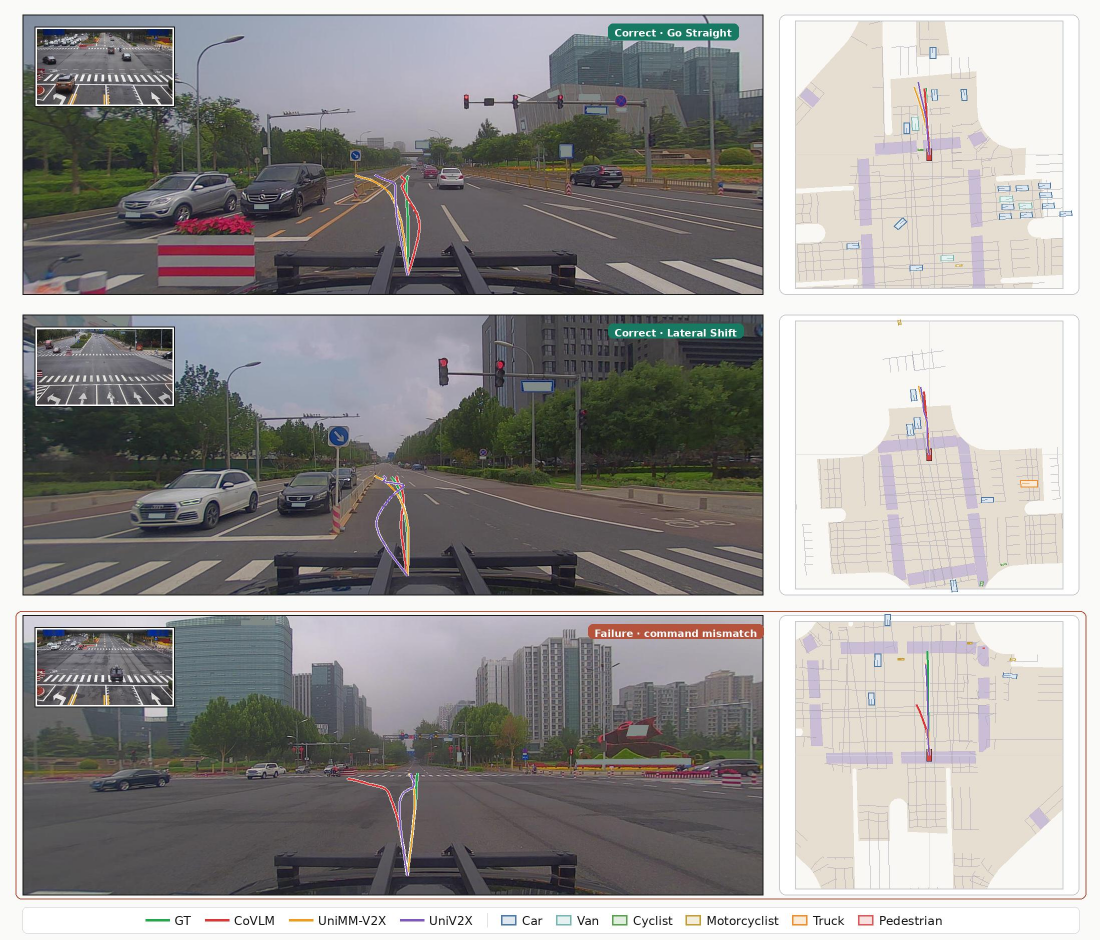}
\captionsetup{type=figure,hypcap=false}
\caption{\textbf{CP examples}. The first two rows show command-correct
straight-driving and lateral-shift cases in which \model{} remains closest to
the reference trajectory among the shown planners. The bottom row shows a
left-turn prediction for a straight-driving reference. Camera projections and
BEV overlays visualize the structured command and waypoint outputs evaluated
in Table~\ref{tab:external-baselines}.}
\label{fig:appendix-waypoint-qualitative}
\end{minipage}
\end{center}

\FloatBarrier

\section{Roadside Image Transmission Cost}
\label{sec:appendix-communication}

The latencies in Table~\ref{tab:external-baselines} measure computation. This
section estimates what the roadside input adds over the
infrastructure-to-vehicle link. Only the roadside image crosses that link: the
ego image is captured on board, and \model{} predicts commands and trajectories
from the two images without exchanging intermediate representations.

Over the 2,196 benchmark frames the stored roadside images are
$1920\times1080$ JPEG averaging 307.9~KiB, or 2.52~Mbit. The reference planner
resizes each image to at most 262,144 pixels before encoding it, so a roadside
node need not transmit more than that resolution; re-encoding at the resulting
$682\times384$ gives 0.42--0.99~Mbit per frame for JPEG quality 75 to 95.
Table~\ref{tab:communication-cost} converts both payloads into transmission
time.

\begin{table}[!htbp]
\caption{\textbf{Roadside image payload and transmission time}, measured over the
2,196 benchmark frames. Cells are transmission time in milliseconds, computed
as mean payload divided by uplink rate. Source images retain their stored JPEG
encoding; resized images are re-encoded at the indicated JPEG quality.}
\label{tab:communication-cost}
\centering
\small
\setlength{\tabcolsep}{5pt}
\renewcommand{\arraystretch}{1.06}
\begin{adjustbox}{max width=\linewidth}
\begin{tabular}{lrrrrr}
\toprule
& \multicolumn{5}{c}{Uplink rate (Mbit/s)} \\
\cmidrule(lr){2-6}
Roadside payload & 10 & 25 & 50 & 100 & 200 \\
\midrule
Source 1920$\times$1080 (2.52~Mbit) & 252.3 & 100.9 & 50.5 & 25.2 & 12.6 \\
Consumed 682$\times$384, quality 95 (0.99~Mbit) & 98.9 & 39.6 & 19.8 & 9.9 & 4.9 \\
Consumed 682$\times$384, quality 85 (0.56~Mbit) & 55.8 & 22.3 & 11.2 & 5.6 & 2.8 \\
Consumed 682$\times$384, quality 75 (0.42~Mbit) & 41.9 & 16.8 & 8.4 & 4.2 & 2.1 \\
\bottomrule
\end{tabular}

\end{adjustbox}
\end{table}

At 10~Mbit/s, a source-resolution image requires 252.3~ms of transmission,
while resized images require 41.9--98.9~ms across the tested JPEG qualities.
These estimates describe the bandwidth cost of the paired-image interface.
Throughput varies with band, load, and mobility
\citep{fezeu2023midband,rochman2023eval}, motivating the range of link rates
reported here. The calculation assumes one roadside image per planning step
and excludes protocol overhead, queueing, retransmission, packet loss, and
contention. It is a payload-based estimate, separate from the measured
computation times in Table~\ref{tab:external-baselines}.

\FloatBarrier

\section{Release, Ethics, and Limitations}
\label{sec:appendix-scope}

\paragraph{Release contents.}
The benchmark package is organized into fixed train/validation splits, L3 question--answer
and L4 rationale annotations, prompts, evaluation scripts, figure and table
scripts, baseline configurations, model metadata, and structured predictions.
It also contains the shuffled-control assignments and predictions,
per-condition evaluation files, metric summaries, and common-device runtime
profiles needed to reproduce the reported controls.
Underlying images and cooperative labels follow the upstream
V2X-Seq-SPD/DAIR-V2X-Seq access terms; users obtain restricted source data
separately where required by those terms.

\paragraph{Ethics and release constraints.}
\dataset{} is derived from traffic-scene sensor records and introduces no
personal identity labels. Released materials follow the privacy filtering and
redistribution constraints of the upstream datasets.
\subsection{Evaluation scope}
The benchmark evaluates CDQA and open-loop trajectory prediction from paired
camera observations. Input controls measure dependence on roadside evidence;
driving safety would additionally require interactive or closed-loop evaluation.

Maneuver QA carries a further limit. Its question wording, reproduced from the
inference prompt, asks for the safest ego action, whereas its reference answer
is the maneuver the ego vehicle executed, discretized from the recorded
trajectory. Accordingly, this metric measures agreement with recorded behavior,
not safety optimality. The prompt is reproduced verbatim to document the
evaluated task.
Because maneuver QA supplies 654 of the 2,861 scored items, we also report the
aggregate over the remaining 2,207 evidence-grounded items
(Table~\ref{tab:appendix-cdqa-conditions}). This separates object-related QA
performance from recorded-maneuver agreement.

The data come from one real-world V2I source and support camera-only, offline
evaluation. Temporal evidence accumulation, calibration and pose errors,
communication impairments, interactive traffic responses, and cross-domain
generalization remain outside the reported tests.
Appendix~\ref{sec:appendix-communication} estimates the transmission payload
of the roadside input under specified link rates.

\FloatBarrier

\end{document}